\documentclass{article}
\PassOptionsToPackage{numbers,compress,sort}{natbib}

\usepackage[preprint]{neurips_2026}

\usepackage[utf8]{inputenc} 
\usepackage[T1]{fontenc}    
\usepackage{hyperref}       
\usepackage{url}            
\usepackage{booktabs}       
\usepackage{amsfonts}       
\usepackage{nicefrac}       
\usepackage{microtype}      
\usepackage[table]{xcolor}  
\definecolor{bestcell}{RGB}{180, 210, 240}   
\definecolor{secondcell}{RGB}{220, 232, 246} 
\usepackage{enumitem}
\usepackage{graphicx}
\usepackage{wrapfig}
\usepackage{pifont}
\usepackage{makecell}
\usepackage{amssymb}
\usepackage{stackengine}
\usepackage{amsmath}
\usepackage{algorithm}
\usepackage{algorithmic}
\usepackage{multirow}
\usepackage[most]{tcolorbox}

\newcommand{\halfright}{\stackon[-3pt]{\ding{51}}{\rule{0.6em}{1pt}}}

\tcbset{
    promptstyle/.style={
        enhanced jigsaw,
        breakable,
        colback=gray!8,
        colframe=black!55,
        boxrule=0.8pt,
        arc=3pt,
        left=8pt, right=8pt, top=6pt, bottom=6pt,
        before skip=10pt, after skip=10pt,
        attach boxed title to top left={yshift=-2.5mm, xshift=3mm},
        fonttitle=\bfseries\small,
        colbacktitle=black!65,
        coltitle=white,
        boxed title style={
            arc=2pt,
            boxrule=0pt,
        },
        before upper={\setlength{\parindent}{0pt}\setlength{\parskip}{0.6\baselineskip}},
    }
}

\title{VisInteract: Towards Dynamic Interactive Text-to-Visualization under Imperfect Queries}

\author{%
  \textbf{Wenxin Xu}\textsuperscript{1}\thanks{Equal contribution.} \quad
  \textbf{Jinwei Lu}\textsuperscript{1}\footnotemark[1] \quad
  \textbf{Hwanhee Kim}\textsuperscript{1} \quad
  \textbf{Chen Jason Zhang}\textsuperscript{1} \\
  \textbf{Xiao-Yong Wei}\textsuperscript{1} \quad
  \textbf{Haoyang Li}\textsuperscript{1} \quad
  \textbf{Yuanfeng Song}\textsuperscript{2} \\[0.6ex]
  \normalfont
  \textsuperscript{1}The Hong Kong Polytechnic University \quad
  \textsuperscript{2}ByteDance \\
}

\begin{document}

\maketitle

\begin{abstract}
Real-world visualization requests are routinely ambiguous, incomplete, or factually incorrect, yet existing Text-to-Visualization (Text-to-Vis) systems assume well-specified inputs and produce charts in a single pass.
When queries are imperfect, a system must \emph{interact} with the user to recover the true intent, but no benchmark or method supports this dynamic process.
We introduce \textbf{VisInteract}, a new paradigm that reframes Text-to-Vis as interaction-driven intent recovery, and \textbf{VisInteract-Bench}, to our knowledge, that is the first benchmark for dynamic interactive Text-to-Vis, featuring controlled imperfection injection, a leakage-controlled User Agent for realistic multi-turn feedback, and dual-perspective (code and chart) automated evaluation.
On the algorithmic side, we propose \textbf{Vis-MCTS}, a Monte Carlo Tree Search (MCTS) enhanced method, introducing improvements over classical MCTS, that \emph{Progressive Widening} to tame the unbounded tool-argument space in tree search, \emph{cross-rollout information sharing} so clarifications and critiques benefit the entire search tree, and \emph{Dimension-Aware Reward Decomposition} that routes scalar user feedback along data-fidelity, visual-design, and intent-alignment dimensions to resolve credit assignment across heterogeneous actions.
Extensive Experiments across two LLM backbones show that Vis-MCTS consistently outperforms all Text-to-Vis baselines, improving end-to-end task success by $13.40\%$--$16.27\%$ over the strongest interactive baseline and by more than $5\times$ over non-interactive ones.
\end{abstract}

\section{Introduction}
\label{sec:intro}

Consider an analyst who types \textit{``Plot the monthly revenue of our top-5 products in 2025, broken down by region.''}
The request looks well-specified, yet \emph{top-5} (by revenue or units?), whether refunds count, and even whether the database covers 2025 are all left implicit. Such under-specification is the rule rather than the exception in real Text-to-Vis usage~\cite{srinivasan2021collecting, narechania2020nl4dv}.
Without a way to ask, a single-pass system silently commits to one interpretation, producing a chart misaligned with the user's true intent.

\begin{figure}[t]
    \centering
    \includegraphics[width=\textwidth]{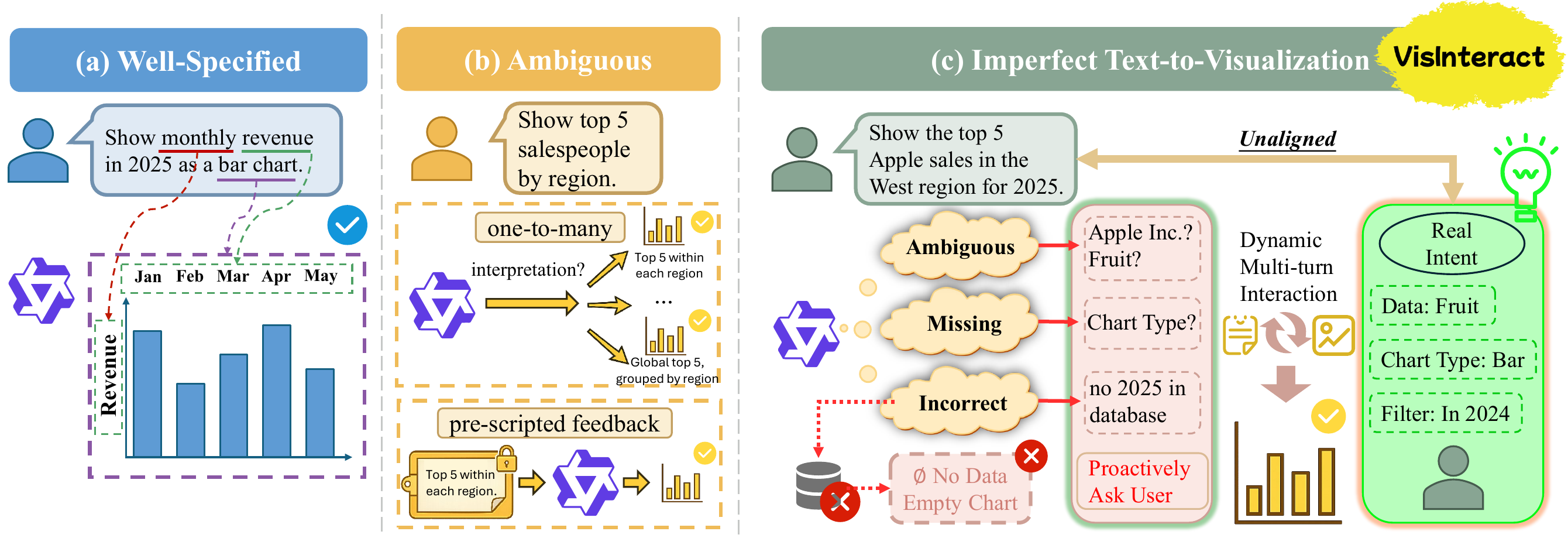}
    \caption{Comparison of  Text-to-Vis paradigms. \textbf{(a)}~Well-specified queries map to a single target chart. \textbf{(b)}~Ambiguous queries admit multiple interpretations, that prior work uses static mappings or pre-scripted feedback. \textbf{(c)}~\textbf{VisInteract} recovers user intent from imperfect queries through dynamic, multi-turn, multi-modal interaction.}
    \label{fig:intro}
\end{figure}

Text-to-Vis has advanced rapidly. On the \emph{method} side, the field has progressed from sequence-to-sequence models targeting declarative grammars such as Vega-Lite or VQL~\cite{luo2021natural}, to LLM-based code generation~\cite{tian2024chartgpt, dibia2023lida, maddigan2023chat2vis}, and most recently to agent-based pipelines~\cite{lu2026multivis, ouyang2025nvagent, xiong2025interactive}.
On the \emph{benchmark} side, NVBench~\cite{luo2021synthesizing} established the first large-scale NL-to-Vis benchmark, with subsequent work expanding through human-authored utterances~\cite{srinivasan2021collecting}, LLM-generated query-chart pairs~\cite{ko2024natural, chen2024viseval}, and multi-chart reasoning~\cite{lu2026multivis}.
Yet all assume the query already faithfully encodes the user's true intent, with systems either rendering in a fixed pipeline or merely reacting to user-initiated edits, never proactively probing what is missing, ambiguous, or incorrect (Figure~\ref{fig:intro}a, b). Once this assumption is relaxed, two fundamental challenges arise.

\textbf{Challenge 1: Existing benchmarks neither model imperfect inputs nor evaluate open-ended outputs.}
On the input side, real-world queries are routinely \emph{ambiguous}, \emph{incomplete}, or \emph{factually incorrect}, yet current benchmarks remain single-turn~\cite{luo2021synthesizing, luo2025nvbench}, pre-scripted~\cite{song2024marrying}, or limited to static feedback~\cite{xiong2025interactive} (Figure~\ref{fig:intro}b). Adjacent NL2SQL benchmarks such as BIRD-INTERACT~\cite{huobird} demonstrate the value of dynamic clarification, but no analogous benchmark exists for visualization, which further requires \emph{visual} feedback on chart design.
On the output side, unlike NL2SQL where result-set equivalence yields a clean correctness signal, visualization is inherently open-ended, and exact-match or execution-equivalence metrics~\cite{luo2021synthesizing, lu2025towards, li2025prompt4vis} penalize correct-but-different designs, conflating \emph{whether} user requirements are met with \emph{how} they are implemented.

\textbf{Challenge 2: Existing methods cannot proactively explore users' true intent through dynamic interactions.}
Pipeline approaches~\cite{tian2024chartgpt, dibia2023lida, lu2026multivis} render a single visualization without any user interaction, ambiguity-focused methods~\cite{luo2025nvbench} merely enumerate a fixed candidate set without interaction, and feedback-based variants~\cite{xiong2025interactive} react only to pre-scripted, user-initiated edits along one trajectory. None can recover from an imperfect initial hypothesis or systematically explore the design space, ultimately failing to converge on the user's true intent.

These gaps motivate \textbf{VisInteract}, a paradigm that reframes Text-to-Vis as dynamic, interaction-driven intent recovery (Figure~\ref{fig:intro}c), with the following contributions:

\begin{itemize}[leftmargin=12pt, itemsep=2pt]
    \item \textbf{VisInteract-Bench} (\S\ref{sec:benchmark}), to our knowledge the first benchmark for dynamic interactive Text-to-Vis, combining \emph{controlled imperfection injection} of ambiguity, incompleteness, and factual errors, a \emph{leakage-controlled User Agent} that simulates realistic multi-turn feedback, and a \emph{dual-perspective evaluation} in which an LLM-as-judge~\cite{zheng2023judging, kim2024prometheus} scores both the generated code and the rendered chart against user intent, avoiding over-penalization by exact-match metrics.
    \item \textbf{Vis-MCTS} (\S\ref{sec:method}), a Monte Carlo Tree Search (MCTS) enhanced method for interactive intent recovery. Recovering intent demands backtracking from early misinterpretations and systematic exploration of alternative designs, which single-trajectory methods cannot provide but MCTS supports natively via branching expansion and feedback-driven rollouts. Yet vanilla MCTS faces three non-trivial obstacles here: each LLM tool-call admits unbounded textual arguments (SQL queries, code snippets, or natural-language questions), making the per-node action space combinatorially infinite and defeating vanilla MCTS's exhaustive expansion; costly user clarifications gained on one rollout are wasted by other trajectories that explore in isolation; and a single scalar reward uniformly backpropagated mis-routes feedback across heterogeneous actions, e.g., a visual-design flaw can penalize a correct data query. We resolve each with a tailored innovation: \emph{Progressive Widening} adaptively samples promising arguments to constrain the action space, \emph{cross-rollout information sharing} propagates clarifications and critiques so every interaction benefits the whole tree, and \emph{Dimension-Aware Reward Decomposition} splits the scalar feedback into data, visual, and intent exploration rewards, so each action node updates only on the dimension it governs.
    \item \textbf{Empirical findings} (\S\ref{sec:experiments}), where Vis-MCTS consistently outperforms strong ReAct~\cite{yao2022react} and multi-agent~\cite{lu2026multivis, xiong2025interactive} baselines, and ablations confirm that proactive clarification, multi-trajectory exploration, and dimension-aware reward decomposition are each essential.
\end{itemize}

\section{Task Definition}
\label{sec:problem}

Unlike SQL, where a query has a unique result set, visualization admits multiple valid designs for the same analytical intent.
We therefore represent user intent not as a single target chart, but as a set of \emph{key features} that any satisfactory visualization must exhibit.

\textbf{Key features.}
Let $\mathcal{K} = \{k_1, \ldots, k_m\}$ denote the user's latent visualization intent, where each key feature $k_i = (\rho_i, d_i, \mu_i)$ pairs a type $\rho_i \in \mathcal{P}$ with a natural-language description $d_i$ and a binary \texttt{must} flag $\mu_i \in \{0, 1\}$ indicating whether satisfying $k_i$ is mandatory ($\mu_i\!=\!1$) or merely preferred ($\mu_i\!=\!0$); we write $\mathcal{K}^{\text{must}} = \{k_i \in \mathcal{K} : \mu_i = 1\}$ for the must-have subset.
The type vocabulary
$\mathcal{P} = \{\texttt{mark},\, \texttt{encoding},\, \texttt{filter},\, \texttt{aggregation},\, \texttt{composition},\, \texttt{interaction}\}$
follows the canonical primitives of the Grammar of Interactive Graphics~\cite{satyanarayan2016vega}, jointly spanning data-level transformations and visual-level specifications, with overlays such as statistical reference lines subsumed under \texttt{composition}. Together these six categories already cover the design decisions exercised by mainstream Text-to-Vis benchmarks~\cite{luo2025nvbench, chen2024viseval}.

\textbf{Traditional Text-to-Vis.}
Given a database $\mathcal{D}$ and a well-specified query $q$ aligned with $\mathcal{K}$, the task is a mapping $f\colon (\mathcal{D}, q) \mapsto \mathcal{Y}$ producing visualization $\mathcal{Y} = (c, z)$, where $c$ is executable code and $z$ is the corresponding rendered chart, such that $\mathcal{Y} \models \mathcal{K}$ (or, in the relaxed sense, $\mathcal{Y} \models \mathcal{K}^{\text{must}}$), that is, both the code $c$ and the rendered chart $z$ satisfy every key feature in $\mathcal{K}$.

\textbf{VisInteract setting.}
In practice, the system receives an imperfect query $\tilde{q} = \pi(\mathcal{K})$, where projection $\pi$ may omit, distort, or under-specify parts of $\mathcal{K}$.
Since $\pi$ is not injective, $\tilde{q}$ is consistent with multiple plausible key-feature sets, making intent convergence through interaction necessary.
We use $\Omega$ to denote the observation process, that the system can obtain indirect evidence about $\mathcal{K}$ through textual clarification or visual feedback, but it never observes $\mathcal{K}$ itself.
VisInteract defines:
\[
g^{\Omega}\colon (\mathcal{D},\, \tilde{q}) \;\mapsto\; \mathcal{Y} \quad \text{s.t.}\quad \mathcal{Y} \models \mathcal{K}.
\]
Since $\mathcal{K}$ is never directly accessible, the system must decide \emph{when} and \emph{what} to ask through $\Omega$, giving rise to a partially observable sequential decision problem (\S\ref{sec:method}).

\begin{table}[t]
\centering
\caption{Comparison of Text-to-Vis benchmarks. To our knowledge, VisInteract-Bench is the first to combine dynamic interaction, multi-modal feedback, and explicitly imperfect queries.}
\label{tab:benchmark-comparison}
\resizebox{\textwidth}{!}{%
\begin{tabular}{lcccccccc}
\toprule
\textbf{Dataset} & \textbf{\#Tables} & \textbf{\#Samples} & \makecell{\textbf{Interaction}\\ \textbf{Type}} & \makecell{\textbf{Interaction}\\ \textbf{Modality}} & \makecell{\textbf{NL Query}\\ \textbf{Imperfect}} & \textbf{Output Format} & \textbf{Construction} \\
\midrule
NLV Corpus \cite{srinivasan2021collecting}    & 3     & 814     & \ding{55} & \ding{55} & \halfright & Vega-Lite   & Human     \\
VL2NL \cite{ko2024natural}           & 1,981 & 3,962   & \ding{55} & \ding{55} & \halfright & Vega-Lite   & LLM       \\
VisEval \cite{chen2024viseval}        & 748   & 2,524   & \ding{55} & \ding{55} & \ding{55} & VQL         & LLM       \\
NVBench \cite{luo2021synthesizing}          & 780   & 25,750  & \ding{55} & \ding{55} & \ding{55} & VQL         & Program   \\
NVBench 2.0 \cite{luo2025nvbench}     & 780   & 24,076  & \ding{55} & \ding{55} & \halfright & VQL         & LLM       \\
MultiVis-Bench \cite{lu2026multivis}  & 697   & 1,202   & \ding{55} & \ding{55} & \ding{55} & Python Code & Human+LLM \\
Dial-NVBench \cite{song2024marrying}    & 780   & 124,449 & Static & Text & \ding{55} & VQL         & Program   \\
NVBench-Feedback \cite{xiong2025interactive}  & -   & 48,482   & Static & Text & \ding{55} & VQL & LLM \\
\midrule
\textbf{VisInteract-Bench} & 75 & 1,098 & Dynamic & Text + Visual & \ding{51} & Python Code & Human+LLM \\
\bottomrule
\end{tabular}%
}
\end{table}

\section{VisInteract-Bench}
\label{sec:benchmark}

Table~\ref{tab:benchmark-comparison} summarizes the Text-to-Vis benchmark landscape. Prior datasets either assume well-specified inputs or offer only static interactions, and none combines dynamic multi-modal interaction, and imperfect queries.
VisInteract-Bench fills this gap with 1{,}098 samples across 11 databases ($75$ tables in total) drawn from BIRD Mini-Dev~\cite{li2023can}.

\textbf{Sample structure.}
As illustrated in Figure~\ref{fig:sample} (Appendix~\ref{app:construction}), each sample is a tuple $(\tilde{q},\, \mathcal{K},\, \mathcal{Y})$ comprising the imperfect query $\tilde{q}$ exposed to the system, the latent key feature set $\mathcal{K}$, and the validated ground-truth visualization $\mathcal{Y}$. 
Key features $\mathcal{K}$ serve three roles simultaneously, namely 1) defining the user's true intent and only exposed to the User Agent, 2) specifying the targets for imperfection injection, and 3) providing the criteria for automated evaluation.

\textbf{Construction pipeline.}
A five-stage pipeline builds each sample.
(1)~\emph{Source data}. 11 databases ($75$ tables in total) from BIRD Mini-Dev~\cite{li2023can} spanning diverse domains (sports, entertainment, education, finance, healthcare, etc.) and schema complexities.
(2)~\emph{Candidate generation}. For each source instance, an LLM grounded in parsed schema and SQL semantics, and steered by a global \emph{diversity tracker} that biases it toward under-represented chart families and ambiguity templates, proposes $M\!=\!10$ structured visualization candidates, which a per-instance \emph{diversity maximizer} then prunes to $N\!=\!5$.
(3)~\emph{Ground-truth generation}. For each candidate, the LLM produces chart-aligned SQL and Altair~\cite{vanderplas2018altair} code, executed in a sandbox with bounded repair loops; samples that fail SQL execution, chart contracts, or visual quality are discarded, and structured plus natural-language key features are extracted from the validated reference.
(4)~\emph{Controlled imperfection injection}. An LLM rewrites the clear query by combining 1--3 of $14$ \emph{ambiguity templates} from three categories (\emph{ambiguity} for multiple plausible readings, \emph{incompleteness} for critical details dropped, and \emph{factual error} for things like a non-existent column), and records the targeted Vega-Lite paths in an ambiguity profile. Crucially, injection occurs after validation, so every imperfect query retains a correct executable reference.
(5)~\emph{Quality control}. Each candidate passes three automated checks, namely \emph{data shape} (SQL result fits the chart contract), \emph{spec coverage} (rendered chart matches $\ge\!80\%$ of the structured key features), and \emph{ambiguity strength} (imperfect query is neither too explicit nor too vague against the ambiguity profile), followed by human review on naturalness, visual quality, and key-feature faithfulness. Only \texttt{pass} samples are kept and re-indexed, yielding 1{,}098 samples. 

VisInteract-Bench is a zero-shot evaluation set. The full construction details and statistics are in Appendices~\ref{app:construction} and~\ref{app:stats}, respectively.

\textbf{Leakage-controlled User Agent.}
The User Agent simulates a user who knows their intent ($\mathcal{K}$, ground-truth code and chart) but never proactively reveals it. It exposes two complementary interfaces, guarding by \emph{access gating} (which subset of $\mathcal{K}$ is visible) and a \emph{response policy} (what the model may verbalize) to prevent ground-truth leakage.
\texttt{ask\_user\_text}$(q_{\text{text}}) \to a_{\text{text}}$ answers a textual question after a pre-filtering step in which a scanner LLM labels which key-feature types the question pertains to, and only the 
matched features are exposed to the User Agent. \texttt{ask\_user\_vis}$(c) \to (s, m)$ executes $c$, shows the rendered chart alongside the ground-truth chart to a VLM, and returns feedback $m$ on the most salient visual discrepancy which describes \emph{what is wrong}, not \emph{how to fix}, with a satisfaction score $s \in [1, 10] \cap \mathbb{Z}$, exposing only visual-dimension features. The two are complementary: text resolves data-level ambiguity while vis captures visual perception. Pre-filtering thus confines each answer to the dimension implicated by the call, while a shared \emph{response policy} mandates refusing any direct request for the ground-truth intent. Full prompts are in Appendix~\ref{app:user-agent}.

\textbf{Dual-judge evaluation.}
For each key feature $k \in \mathcal{K}$, two complementary binary LLM-as-Judge~\cite{zheng2023judging, liu2023g} decide whether a submission $\mathcal{Y}=(c, z)$ satisfies $k$. The \emph{Code Judge} $J_\text{code}(c, k) \in \{0,1\}$ checks the predicted code $c$ against $k$ with line-level evidence, while the \emph{Chart Judge} $J_\text{chart}(z, k) \in \{0,1\}$ inspects the rendered chart $z$ to verify $k$ visually, catching code-invisible issues such as overlapping labels or mis-mapped colors. This yields three reporting perspectives:
\begin{equation}
\label{eq:sigma}
\sigma_\text{code}(k) = J_\text{code}(c, k), \quad \sigma_\text{chart}(k) = J_\text{chart}(z, k), \quad \sigma_\text{merge}(k) = \sigma_\text{code}(k) \,\land\, \sigma_\text{chart}(k).
\end{equation}
Over $N$ samples, with $\mathcal{K}_i^\text{must}\!\subseteq\!\mathcal{K}_i$ the must-have subset and $\sigma(\cdot)$ ranging over the three perspectives in Eq.~\ref{eq:sigma}, we report \textbf{KF Pass Rate} $=\!\sum_i\!\sum_{k\in\mathcal{K}_i}\!\sigma(k)/\!\sum_i\!|\mathcal{K}_i|$ (dataset-wide feature satisfaction), \textbf{Strict Success} $=|\{i:\sigma(k)\!=\!1\,\forall k\!\in\!\mathcal{K}_i^\text{must}\}|/N$ (samples passing every must-have feature), \textbf{Task Score} $=|\{i:\sigma(k)\!=\!1\,\forall k\!\in\!\mathcal{K}_i\}|/N$ (samples passing every feature), and \textbf{Renderable Rate} (percentage of submissions whose code executes and renders without error). We further analyze the reliability of our LLM-as-Judge in Section~\ref{sec:exp-analysis}.

\section{Vis-MCTS: Dynamic Interactive Text-to-Vis via Monte Carlo Tree Search}
\label{sec:method}

\begin{figure}[t]
    \centering
    \includegraphics[width=\textwidth]{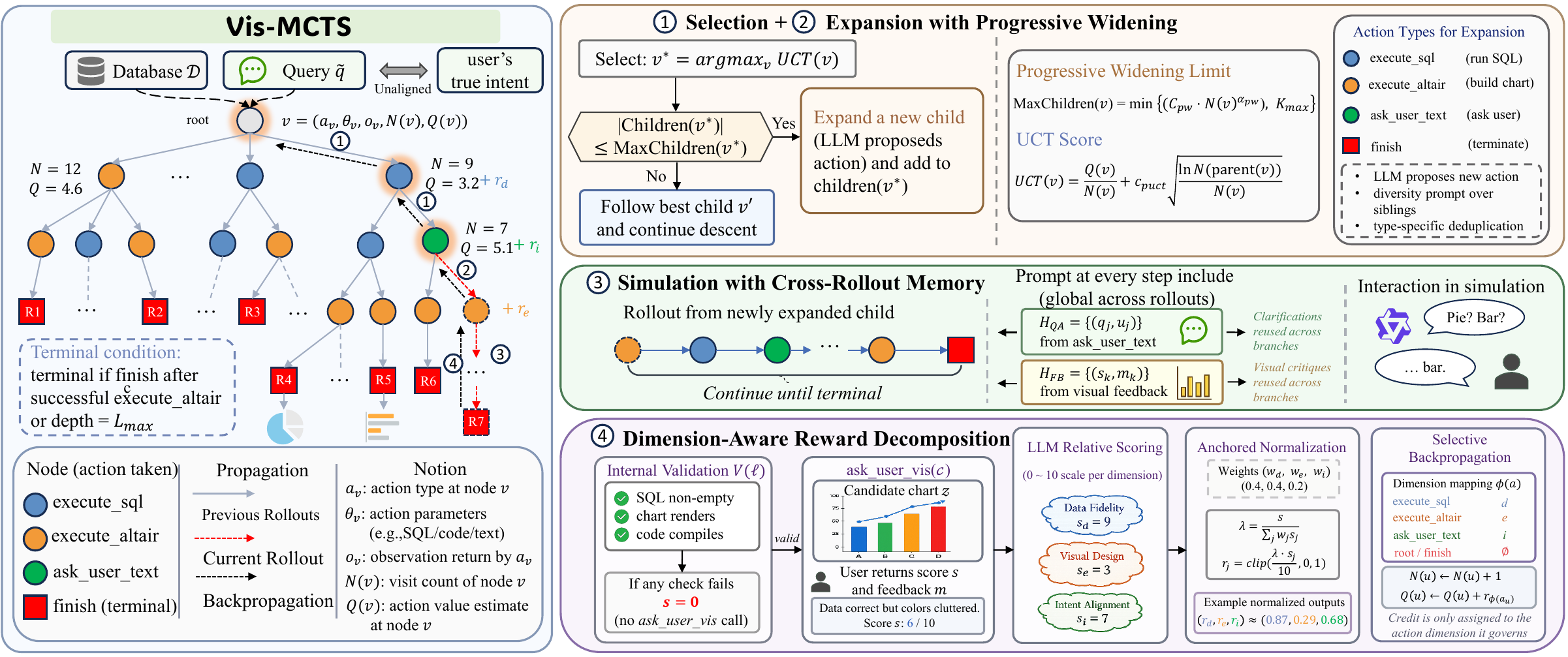}
    \caption{Overview of our proposed Vis-MCTS.}
    \label{fig:vis-mcts}
\end{figure}

\subsection{Overview}
\label{sec:method-overview}

Figure~\ref{fig:vis-mcts} illustrates the complete Vis-MCTS framework.
We model dynamic interactive Text-to-Vis as a partially observable sequential decision problem~\cite{kaelbling1998planning}, that the system observes an imperfect query $\tilde{q}$ and a database $\mathcal{D}$ while the true user intent $\mathcal{K}$ remains latent, and adopt Monte Carlo Tree Search (MCTS)~\cite{browne2012survey} as the underlying search backbone.
We define the LLM agent's action space be $\mathcal{A} = \{\texttt{execute\_sql},\, \texttt{execute\_altair},\, \texttt{ask\_user\_text},\,  \texttt{ask\_user\_vis},\,\texttt{finish}\}$.
At each decision step $t$, the agent selects an action $a_t \in \mathcal{A}$ conditioned on the accumulated history $\{(a_i, o_i)| 0 < i < t\}$, invokes the corresponding tool, and receives an observation $o_t$.
Crucially, \texttt{ask\_user\_vis} is not a selectable action during tree search exploration, that the search framework invokes it only at each valid terminal node to obtain a user satisfaction score $s$ from the User Agent, which serves as the reward signal for backpropagation and the objective that Vis-MCTS maximizes.

Adapting traditional MCTS to interactive Text-to-Vis raises three challenges that existing LLM tree-search methods~\cite{yao2023tree, zhou2024language} leave unaddressed: an unbounded action parameter space, the lack of cross-rollout information sharing, and incorrect credit assignment under scalar-reward backpropagation.
Vis-MCTS addresses them via \emph{Progressive Widening} (\S\ref{sec:method-pw}), persistent cross-rollout memories during simulation (\S\ref{sec:method-sim}), and \emph{Dimension-Aware Reward Decomposition} (\S\ref{sec:method-decompose}), respectively.

\subsection{Search Tree Structure}
\label{sec:method-tree}

Each node $v$ in the search tree corresponds to one tool-call step in a ReAct-style~\cite{yao2022react} reasoning chain, formally represented as:
\begin{equation}
\label{eq:node}
v = \bigl(a_v,\, \theta_v,\, o_v,\, N(v),\, Q(v)\bigr),
\end{equation}
where $a_v \in \mathcal{A}$ is the action, $\theta_v$ is its parameters (e.g., SQL text or Altair code), $o_v$ is the tool observation, $N(v) \in \mathbb{N}$ is the \emph{visit count} that counts how many rollouts have traversed $v$, and $Q(v) \in \mathbb{R}_{\ge 0}$ is the \emph{cumulative reward} backpropagated through $v$ so far.
A root-to-leaf path $\tau = (v_0, v_1, \ldots, v_T)$ constitutes a complete LLM reasoning trajectory, that is a candidate solution.
Children of the same parent represent alternative actions explored at that decision point.

A node $v$ is \emph{terminal} if $a_v = \texttt{finish}$ and the most recent \texttt{execute\_altair} ancestor succeeded, or if the path depth reaches $L_\text{max}$.
Terminal nodes that pass an internal validation predicate $\mathcal{V}\colon v \to \{0,1\}$, which verifying that code compiles, the chart renders, and SQL returns non-empty results, are submitted for user evaluation; those with $\mathcal{V}(v) = 0$ receive zero reward without invoking \texttt{ask\_user\_vis} for further user feedback.

\subsection{Selection with Progressive Widening}
\label{sec:method-pw}

In classical MCTS, selection descends the tree by choosing the child with the highest Upper Confidence bound for Trees (UCT)~\cite{browne2012survey} score, then \emph{exhaustively} expands all legal moves at the selected node.
This is feasible when the action space is discrete and small (e.g., board games), but in Vis-MCTS the parameter space $\Theta_a$ of each action is effectively unbounded, where the LLM can generate infinitely many distinct SQL queries, Altair programs, or clarification questions, making exhaustive expansion infeasible.

We therefore adopt \emph{Progressive Widening}~\cite{couetoux2011continuous} (Figure~\ref{fig:vis-mcts}, \ding{172}\ding{173}), which caps the number of children at each node and lets it grow sublinearly with the visit count:
\begin{equation}
\label{eq:pw}
\text{MaxChildren}(v) = \min\!\bigl(\lceil C_\text{pw} \cdot N(v)^{\alpha_\text{pw}} \rceil,\; K_\text{max}\bigr),
\end{equation}
where $C_\text{pw}$ controls the widening rate, $\alpha_\text{pw}$ governs growth exponent, and $K_\text{max}$ is a hard cap.
As a result, selection no longer relies on UCT alone but jointly considers UCT and $\text{MaxChildren}$. At each node $v$ during descent, if $|\text{children}(v)| \leq \text{MaxChildren}(v)$, the node is \emph{expandable} and a new child is expanded; otherwise its child with the highest UCT score is selected and descent continues via
\begin{equation}
\label{eq:uct}
\text{UCT}(v) = \underbrace{\frac{Q(v)}{N(v)}}_{\text{exploitation}} + \underbrace{c_\text{puct}\sqrt{\frac{\ln N(\text{parent}(v))}{N(v)}}}_{\text{exploration}},
\end{equation}
where $c_\text{puct} > 0$ is an exploration constant that balances the two terms. The first favors children with high average reward, while the second encourages visiting under-explored branches.

When a node is expandable, the framework prompts the LLM for a new action within the legal subspace defined by a successor function $\Gamma\colon \mathcal{A} \to 2^{\mathcal{A}}$ that maps each action to its permissible successors (Appendix~\ref{app:successor}).
Since the LLM tends to produce similar outputs under the same context, a \emph{diversity prompt} describes existing sibling actions to encourage novelty (Appendix~\ref{app:diversity}), and a type-specific \emph{deduplication check} discards any candidate whose action type and canonicalized key argument exactly match an existing sibling (Appendix~\ref{app:dedup}).
The canonicalization uses trimmed and lowercased text for SQL and question actions, and a fixed-length prefix for Altair code.
If expanding succeeds, the new child is passed to simulation; if rejected by deduplication, selection falls back to UCT descent and retries at a deeper node.
This process repeats at each level until a new node is successfully created or the tree is exhausted.

\subsection{Simulation}
\label{sec:method-sim}

After expanding produces a new node, simulation extends the path to a terminal state (Figure~\ref{fig:vis-mcts}, \ding{174}) by repeatedly invoking LLM for the next action, subject to the successor constraints $\Gamma$, until a \texttt{finish} node is reached.
Throughout this process, two global memory structures are injected into the LLM prompt to enable \emph{cross-rollout information sharing}, namely (i) a QA history $\mathcal{H}_\text{QA} = \{(q_j, u_j)\}$ that accumulates all \texttt{ask\_user\_text} exchanges, where $q_j$ is the clarification question and $u_j$ the user's reply, and (ii) a feedback history $\mathcal{H}_\text{FB} = \{(s_k, m_k)\}$ that records all \texttt{ask\_user\_vis} results, where $s_k \in [1, 10] \cap \mathbb{Z}$ is the satisfaction score and $m_k$ the textual critique.
Since both histories are shared across \emph{all} rollouts, a single interaction benefits the entire tree, not just the branch that generated it.

\subsection{Dimension-Aware Reward Decomposition}
\label{sec:method-decompose}

Once simulation reaches a terminal node $\ell$, Vis-MCTS obtains user feedback and decomposes it into dimension-aware rewards (Figure~\ref{fig:vis-mcts}, \ding{175}).
First, an internal validation $\mathcal{V}(\ell)$ catches failures (code errors, empty SQL results, render failures) and assigns $s = 0$ without invoking \texttt{ask\_user\_vis}.
If $\mathcal{V}(\ell) = 1$, the search framework calls \texttt{ask\_user\_vis} that the User Agent returns a satisfaction score $s$ and textual feedback $m$, which are appended to $\mathcal{H}_\text{FB}$.

Standard MCTS would backpropagate this scalar $s$ uniformly to all ancestors, but a visualization trajectory interleaves SQL nodes (data retrieval), Altair nodes (chart design), and text-query nodes (intent clarification), creating a credit-assignment problem: when the user reports ``data is correct but the chart type is wrong'' with a low overall score, a SQL node that issued a correct query is unfairly penalized for an unrelated visual deficiency.
Thus we decompose the scalar reward into three interpretable dimensions, each aligned with an action type:

\textbf{Step 1: Dimension-wise relative scoring.}
We define a scoring function $\mathcal{R}\colon \mathcal{I} \times \mathcal{T} \times \mathcal{S} \times \mathcal{M} \;\to\; \mathcal{S}^{3}$, where $\mathcal{I}$ denotes the rendered chart images, $\mathcal{T}$ denotes the trajectory, $\mathcal{S}$ denotes the user scores, and $\mathcal{M}$ denotes the corresponding textual feedback.
An LLM instantiates $\mathcal{R}$ to produce per-dimension scores $(s_d, s_e, s_i) = \mathcal{R}(\text{img},\, \tau,\, s,\, m)$ reflecting data fidelity, visualization design, and intent alignment respectively.
The prompt focuses on \emph{relative} quality differences, which dimensions the feedback praises or criticizes, rather than absolute calibration (full prompt template in Appendix~\ref{app:reward-prompt}).

\textbf{Step 2: Anchored normalization.}
We scale the LLM-assigned scores so that their weighted combination exactly equals the original user score, preserving the total reward injected into the tree:
\begin{equation}
\label{eq:decompose}
\lambda = \frac{s}{\textstyle\sum_{j \in \{d,e,i\}} w_j \cdot s_j}, \qquad
r_j = \operatorname{clip}\!\Bigl(\frac{\lambda \cdot s_j}{10},\; 0,\; 1\Bigr) \;\;\forall\, j \in \{d, e, i\},
\end{equation}
where $(w_d, w_e, w_i)$ are dimension weights with $\sum_j w_j = 1$.
The normalization guarantees $\sum_{j} w_j \cdot r_j = s / 10$, ensuring that the total reward flowing into the tree equals the original user score regardless of the LLM's scoring tendencies. A concrete example is provided in Appendix~\ref{app:case-study}.

\subsection{Selective Backpropagation}
\label{sec:method-backprop}

Given decomposed reward vector $(r_d, r_e, r_i)$, we define a dimension mapping $\phi\colon \mathcal{A} \to \{d, e, i, \emptyset\}$ that sends \texttt{execute\_sql}, \texttt{execute\_altair}, \texttt{ask\_user\_text} to $d, e, i$ respectively, and \texttt{finish}/\texttt{root} to $\emptyset$. During backpropagation from terminal leaf $\ell$, each ancestor $u$ updates:
\begin{equation}
\label{eq:backprop}
N(u) \leftarrow N(u) + 1, \qquad Q(u) \leftarrow Q(u) + r_{\phi(a_u)} \cdot \mathbf{1}[\phi(a_u) \neq \emptyset].
\end{equation}
Except for \texttt{root} and \texttt{finish}, which update only $N$, each node accumulates a $Q$-value drawn solely from the dimension it governs, namely $r_d$ at \texttt{execute\_sql}, $r_e$ at \texttt{execute\_altair}, and $r_i$ at \texttt{ask\_user\_text}.
The exploitation term $Q(v)/N(v)$ therefore reflects the node's own-dimension performance rather than a mixed average across dimensions, making useful partial trajectories less likely to be penalized for errors made in unrelated later decisions.
A node is treated as exhausted only if it cannot admit new children under the current progressive-widening cap and all existing children are exhausted. Because $\text{MaxChildren}(v)$ increases with $N(v)$, this condition is re-evaluated as visit counts change, preventing early saturation from permanently closing a branch.

The complete Vis-MCTS procedure is summarized in Algorithm~\ref{alg:vis-mcts} in Appendix~\ref{app:algorithm}.

\section{Experiments}
\label{sec:experiments}

\subsection{Experimental Setup}
\label{sec:exp-setup}

\textbf{Benchmark and evaluation.}
We evaluate on the full VisInteract-Bench with 1,098 samples across 11 databases.
All methods are assessed by the dual-judge pipeline (\S\ref{sec:benchmark}), with \texttt{Qwen3.5-plus} serving as the judge LLM. We report KF Pass Rate, Strict Success, and Task Score under Code, Chart, and Merge perspectives, with every reported number averaged over three independent evaluation runs.

\label{sec:exp-main}

\begin{table}[t]
\centering
\caption{Main results on VisInteract-Bench. $^\dagger$Non-interactive methods. $^\ddagger$Interactive methods. Within each backbone, the best result is \colorbox{bestcell}{\textbf{highlighted in blue}} and the second-best in \colorbox{secondcell}{\textbf{light blue}}.}
\label{tab:main-results}
\resizebox{\textwidth}{!}{%
\begin{tabular}{l ccc ccc ccc c}
\toprule
& \multicolumn{3}{c}{\textbf{Code-level}} & \multicolumn{3}{c}{\textbf{Chart-level}} & \multicolumn{3}{c}{\textbf{Merge}} & \\
\cmidrule(lr){2-4} \cmidrule(lr){5-7} \cmidrule(lr){8-10}
\textbf{Method} & KF Score & Strict & Task & KF Score & Strict & Task & KF Score & Strict & Task & Renderable \\
\midrule
\multicolumn{11}{c}{\textbf{\textit{Qwen3.5-flash}}} \\
\midrule
Self-Correction LLM$^\dagger$ & 37.19 & 38.43 & 11.02 & 24.71 & 30.28 & 10.11 & 21.76 & 26.84 &  7.38 & 95.54 \\
nvAgent$^\dagger$             & 11.68 & 14.57 &  2.19 &  9.81 & 12.02 &  2.19 &  8.08 & 10.12 &  1.64 & 66.67 \\
MultiVis-Agent$^\ddagger$     & 41.67 & 43.03 & 15.19 & 29.00 & 35.47 & 12.71 & 25.97 & 32.07 &  9.48 & 94.35 \\
ReAct$^\ddagger$              & 66.62 & 69.72 & 41.17 & 56.25 & 66.10 & 38.80 & 50.83 & 60.38 & 29.05 & 94.72 \\
Best-of-N (ReAct)$^\ddagger$  & \cellcolor{secondcell}\textbf{74.22} & \cellcolor{secondcell}\textbf{77.64} & \cellcolor{secondcell}\textbf{50.00} & \cellcolor{secondcell}\textbf{64.35} & \cellcolor{secondcell}\textbf{75.36} & \cellcolor{secondcell}\textbf{46.81} & \cellcolor{secondcell}\textbf{59.68} & \cellcolor{secondcell}\textbf{70.40} & \cellcolor{secondcell}\textbf{38.06} & \cellcolor{secondcell}\textbf{99.73} \\
\midrule
Vis-MCTS         & \cellcolor{bestcell}\textbf{81.06} & \cellcolor{bestcell}\textbf{84.92} & \cellcolor{bestcell}\textbf{61.66} & \cellcolor{bestcell}\textbf{73.87} & \cellcolor{bestcell}\textbf{85.96} & \cellcolor{bestcell}\textbf{64.39} & \cellcolor{bestcell}\textbf{69.11} & \cellcolor{bestcell}\textbf{81.05} & \cellcolor{bestcell}\textbf{51.46} & \cellcolor{bestcell}\textbf{100} \\
$\triangle$       & $\uparrow$6.84 & $\uparrow$7.28 & $\uparrow$11.66 & $\uparrow$9.52 & $\uparrow$10.60 & $\uparrow$17.58 & $\uparrow$9.43 & $\uparrow$10.65 & $\uparrow$13.40 & $\uparrow$0.27 \\
\midrule
\multicolumn{11}{c}{\textbf{\textit{Gemini-3.1-flash-lite-preview}}} \\
\midrule
Self-Correction LLM$^\dagger$ & 41.29 & 41.59 & 13.93 & 30.00 & 36.38 & 14.21 & 26.00 & 31.85 & 10.66 & 97.63 \\
nvAgent$^\dagger$             & 15.43 & 19.45 &  3.28 & 13.19 & 16.32 &  3.28 & 10.71 & 13.53 &  2.00 & 81.06 \\
MultiVis-Agent$^\ddagger$     & 56.05 & 56.99 & 30.33 & 44.96 & 53.45 & 29.69 & 40.41 & 48.14 & 22.13 & 98.54 \\
ReAct$^\ddagger$              & 68.64 & 71.46 & 47.81 & 60.79 & 71.03 & 47.72 & 56.50 & 66.25 & 39.62 & 95.17 \\
Best-of-N (ReAct)$^\ddagger$  & \cellcolor{secondcell}\textbf{75.43} & \cellcolor{secondcell}\textbf{76.35} & \cellcolor{secondcell}\textbf{54.79} & \cellcolor{secondcell}\textbf{62.92} & \cellcolor{secondcell}\textbf{71.51} & \cellcolor{secondcell}\textbf{53.79} & \cellcolor{secondcell}\textbf{57.90} & \cellcolor{secondcell}\textbf{70.78} & \cellcolor{secondcell}\textbf{40.92} & \cellcolor{secondcell}\textbf{99.73} \\
\midrule
Vis-MCTS         & \cellcolor{bestcell}\textbf{83.06} & \cellcolor{bestcell}\textbf{86.61} & \cellcolor{bestcell}\textbf{67.21} & \cellcolor{bestcell}\textbf{77.32} & \cellcolor{bestcell}\textbf{87.98} & \cellcolor{bestcell}\textbf{69.67} & \cellcolor{bestcell}\textbf{72.77} & \cellcolor{bestcell}\textbf{81.97} & \cellcolor{bestcell}\textbf{57.19} & \cellcolor{bestcell}\textbf{100} \\
$\triangle$       & $\uparrow$7.63 & $\uparrow$10.26 & $\uparrow$12.42 & $\uparrow$14.40 & $\uparrow$16.47 & $\uparrow$15.88 & $\uparrow$14.87 & $\uparrow$11.19 & $\uparrow$16.27 & $\uparrow$0.27 \\
\bottomrule
\end{tabular}
}
\end{table}

\textbf{Baselines.}
We compare Vis-MCTS against non-interactive and interactive baselines.
\emph{Without user interaction}:
\textbf{Self-Correction LLM}, which generates a visualization with LLM and iterative self-correction upon execution errors;
\textbf{nvAgent}~\cite{ouyang2025nvagent}, a fixed workflow with Processor, Composer and Validator originally designed for VQL followed by translating to python code.
\emph{With user interaction}:
\textbf{ReAct}~\cite{yao2022react}, a reasoning-and-acting agent that interleaves chain-of-thought reasoning with tool execution in a single trajectory;
\textbf{Best-of-N (ReAct)}, which runs $N\!=\!10$ independent ReAct rollouts and returns the one with the highest user score;
and \textbf{MultiVis-Agent}~\cite{lu2026multivis}, a multi-agent system with logic rules constraints, consisting of a Coordinator, sql/code Generators and a Validator.
We implement all methods using two backbone models, \texttt{Qwen3.5-flash} and \texttt{Gemini-3.1-flash-lite-preview}, abbreviated as \texttt{Qwen} and \texttt{Gemini} in the following text. More details are in Appendix~\ref{app:baselines}.

\textbf{Vis-MCTS configuration.}
We use $N_\text{rollout} = 10$ rollouts, maximum trajectory depth $L_\text{max} = 20$, and progressive-widening child cap $K_\text{max} = 3$. Full hyperparameter settings are listed in Appendix~\ref{app:hyperparams}.

\subsection{Main Results}

Table~\ref{tab:main-results} presents the main results with the mean std $\approx 0.86$.
We highlight three findings:

\textbf{Vis-MCTS dominates across all perspectives.}
Vis-MCTS attains the best performance on every metric under both backbones, reaching Code/Chart/Merge Task Scores of 61.66\% / 64.39\% / 51.46\% on Qwen and 67.21\% / 69.67\% / 57.19\% on Gemini, with gains of $+11.66\%$ / $+17.58\%$ / $+13.40\%$ and $+12.42\%$ / $+15.88\%$ / $+16.27\%$ over the strongest baseline on each perspective.
The large margin on the jointly demanding Merge perspective, indicating that the gains come from co-optimizing data correctness and visual design rather than a single modality.
Vis-MCTS also achieves a 100\% Renderable rate on both backbones (vs.\ 94.72\% / 95.17\% for ReAct and only 66.67\% / 81.06\% for nvAgent), showing that tree-structured exploration with execution feedback reliably yields valid outputs without trading correctness for coverage.

\textbf{Interaction is essential.}
Both non-interactive baselines, Self-Correction LLM and nvAgent, achieve the lowest scores across all metrics and backbones, even the stronger Self-Correction LLM trails Vis-MCTS by more than $5\times$ on the Merge Task Score (7.38\% vs.\ 51.46\% on Qwen; 10.66\% vs.\ 57.19\% on Gemini).
This large gap quantifies the \emph{Interaction Gain} from dynamic intent recovery, confirming that imperfect queries cannot be reliably resolved by reasoning and self-correction alone.

\textbf{Tree search outperforms single-chain reasoning.}
Under an identical $N\!=\!10$ rollout budget, Vis-MCTS clearly outperforms Best-of-N (ReAct) on Merge Task Score (51.46\% / 57.19\% vs.\ 38.06\% / 40.92\% on Qwen / Gemini, $+13.40\%$ / $+16.27\%$ improvements), while Best-of-N's own gain over a single ReAct chain is smaller and inconsistent across backbones ($+9.01\%$ on Qwen, $+1.30\%$ on Gemini).
This isolates the benefit of the tree structure itself rather than repeated sampling, that Vis-MCTS explores alternative actions at each decision point, reuses successful prefixes, backtracks from dead ends, and shares clarifications and critiques across rollouts.

\subsection{Ablation Studies}
\label{sec:exp-ablation}

To isolate each Vis-MCTS component, we evaluate four variants that disable the two interaction channels (textual and visual) and the two algorithmic contributions that distinguish Vis-MCTS from standard MCTS.
\textbf{w/o Textual Interaction.} Remove \texttt{ask\_user\_text} from the action space, that the agent recovers intent through code execution and visual critiques alone.
\textbf{w/o Visual Feedback.} Disable \texttt{ask\_user\_vis}, that no terminal reward is obtained, so $Q(v)\!\equiv\!0$ and UCT reduces to pure exploration.
\textbf{w/o Cross-Rollout Memory.} Reset $\mathcal{H}_\text{QA}$ and $\mathcal{H}_\text{FB}$ at each rollout so clarifications and critiques do not persist across rollouts.
\textbf{w/o Reward Decomposition.} Replace Eq.~\ref{eq:decompose} with uniform backpropagation, propagating $s/10$ to all ancestors regardless of action type.

\begin{table}[t]
\centering
\caption{Ablation study on Vis-MCTS components with \texttt{Qwen3.5-flash} as the backbone.}
\label{tab:ablation}
\resizebox{\textwidth}{!}{%
\begin{tabular}{l ccc ccc ccc c}
\toprule
& \multicolumn{3}{c}{\textbf{Code-level}} & \multicolumn{3}{c}{\textbf{Chart-level}} & \multicolumn{3}{c}{\textbf{Merge}} & \\
\cmidrule(lr){2-4} \cmidrule(lr){5-7} \cmidrule(lr){8-10}
\textbf{Variant} & KF Score & Strict & Task & KF Score & Strict & Task & KF Score & Strict & Task & \textbf{Renderable} \\
\midrule
Vis-MCTS (full)                     & \textbf{81.06} & \textbf{84.92} & \textbf{61.66} & \textbf{73.87} & \textbf{85.96} & \textbf{64.39} & \textbf{69.11} & \textbf{81.05} & \textbf{51.46} & \textbf{100.00} \\
- w/o Textual Interaction       & 76.24 & 79.69 & 51.82 & 69.09 & 81.05 & 54.64 & 63.55 & 74.93 & 44.08 & 99.82 \\
- w/o Visual Feedback           & 64.58 & 68.63 & 33.42 & 55.31 & 66.20 & 32.79 & 49.99 & 60.26 & 24.24 & 99.73 \\
- w/o Cross-Rollout Memory      & 68.23 & 72.88 & 40.07 & 63.14 & 75.77 & 43.99 & 57.30 & 69.14 & 34.06 & 99.91 \\
- w/o Reward Decomposition      & 73.77 & 77.78 & 50.27 & 68.49 & 77.50 & 51.91 & 61.75 & 71.22 & 43.26 & 99.82 \\
\bottomrule
\end{tabular}%
}
\end{table}

Table~\ref{tab:ablation} confirms every component plays a distinct, non-redundant role.
\emph{Visual Feedback} supplies the terminal reward signal, whose removal collapses UCT to blind exploration, dropping Merge Task Score by $27.22\%$. \emph{Textual Interaction} resolves non-visual intent ambiguities (e.g., filters, aggregations) that visual critiques cannot expose, resulting in $7.38\%$ Merge Task Score decrease if remove it.
\emph{Cross-Rollout Memory} persists $\mathcal{H}_\text{QA}/\mathcal{H}_\text{FB}$ across rollouts; without it, search efficiency degrades, yielding a $17.40\%$ Merge Task Score loss.
\emph{Reward Decomposition} routes per-dimension rewards to the responsible action types. Removing it, uniform backpropagation dilutes joint code--chart attribution and degrades Merge Task Score by $8.20\%$.

\subsection{Analysis}
\label{sec:exp-analysis}

\textbf{Hyperparameter sensitivity.}
The number of rollouts $N_\text{rollout}$ and the maximum trajectory depth $L_\text{max}$ both raise Merge Task Score monotonically with diminishing returns. $N_\text{rollout}$ from 6 to 10 contributes $+11.7\%$ and $L_\text{max}$ from 10 to 20 adds $+6.4\%$, placing the default $(10, 20)$ near the compute--performance knee. A separate sweep over the reward weights $(w_d, w_e, w_i)$ across the uniform allocation and three single-dimension-heavy corners keeps Merge Task Score within $1.64\%$ of the default, confirming that the dimension-aware decomposition does not require fine-tuned weights. Full results of the hyperparameters are in Appendix~\ref{app:hyperparam-sensitivity}.

\textbf{Case study.}
On an imperfect query that bundles a factual error with a visualization mismatch, Vis-MCTS recovers via asking user for clarification, and Dimension-Aware Reward Decomposition selectively penalises only the chart node. The full case is illustrated in Figure~\ref{fig:case-study} in Appendix~\ref{app:case-study}.

\begin{wrapfigure}[10]{r}{0.40\textwidth}
\vspace{-15pt}
\centering
\includegraphics[width=0.40\textwidth]{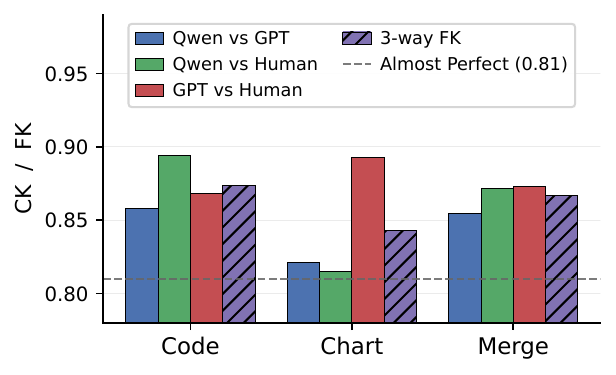}
\vspace{-15pt}
\caption{Inter-judge agreement on VisInteract-Bench. CK denotes Cohen's Kappa coefficient $\kappa$; FK denotes Fleiss' Kappa $\kappa_F$.}
\label{fig:judge-reliability}
\vspace{-12pt}
\end{wrapfigure}
\textbf{Judge reliability.}
To rule out judge bias in Table~\ref{tab:main-results}, we cross-validate our judge LLM (\texttt{Qwen3.5-plus}, abbrev.\ \texttt{Qwen}) against a second judge from a different family (\texttt{GPT-5.4-mini}, abbrev.\ \texttt{GPT}) and human experts on stratified subsets of the benchmark. The detailed settings are deferred to Appendix~\ref{app:judge-reliability}.
As shown in Figure~\ref{fig:judge-reliability}, all pairwise CK and 3-way FK surpass the $0.81$ ``almost perfect'' threshold~\citep{landis1977measurement}.
Crucially, our \texttt{Qwen} judge agrees with humans on par with \texttt{GPT} (mean CK $=\!0.860$ vs.\ $0.878$), so the observed gaps are not artefacts of a single judge family.

\section{Related Work}
\label{sec:related}

\textbf{Text-to-Vis and interactive NL interfaces.}
NVBench~\cite{luo2021synthesizing} established the first large-scale Text-to-Vis benchmark, with subsequent efforts expanding the scope via human-authored utterances~\cite{srinivasan2021collecting}, LLM-generated query--chart pairs~\cite{ko2024natural, chen2024viseval}, and multi-chart reasoning~\cite{lu2026multivis}.
More recent benchmarks shift toward interaction. NVBench~2.0~\cite{luo2025nvbench} addresses query ambiguity but remains single-turn, while Dial-NVBench~\cite{song2024marrying} and NVBench-Feedback~\cite{xiong2025interactive} introduce multi-turn or feedback-based exchanges that are pre-scripted and do not adapt to system output.
Methodologically, approaches have evolved from neural translation~\cite{luo2021natural} to LLM-based step-wise reasoning~\cite{tian2024chartgpt}, code generation~\cite{dibia2023lida, maddigan2023chat2vis}, and concept-driven authoring~\cite{wang2023data}.
In the parallel NL2SQL domain, MISP~\cite{yao2019model} and BIRD-INTERACT~\cite{huobird} pioneered dynamic clarification for resolving ambiguities.
VisInteract-Bench extends this dynamic-clarification paradigm from SQL text to \emph{rendered charts} through controlled imperfection injection and leakage-controlled multi-modal user feedback, evaluating whether a system can recover intent.

\textbf{LLM agents and tree search for reasoning.}
Single-chain agents such as ReAct~\cite{yao2022react}, Reflexion~\cite{shinn2023reflexion}, and CodeAct~\cite{wang2024executable} interleave reasoning with acting but cannot recover from early mistakes.
Tree-structured reasoning addresses this limitation. Tree-of-Thoughts~\cite{yao2023tree} and LATS~\cite{zhou2024language} extend chain-of-thought with search, while RAP~\cite{hao2023reasoning}, MCTSr~\cite{zhang2024accessing}, and rStar~\cite{qi2024rstar} combine MCTS with LLMs for planning and self-refinement.
However, these methods rely on LLM self-evaluation as the reward signal and apply uniform backpropagation.
Vis-MCTS differs in two respects. Progressive Widening~\cite{couetoux2011continuous} handles the open-ended action parameter space of LLM tool-use, and Dimension-Aware Reward Decomposition replaces uniform backpropagation.

\section{Conclusion}
\label{sec:conclusion}

We introduced \textbf{VisInteract}, a paradigm that reframes Text-to-Vis as dynamic, interaction-driven intent recovery from imperfect queries.
VisInteract-Bench provides a benchmark with controlled imperfections, a leakage-controlled User Agent, and dual-perspective automated evaluation.
Vis-MCTS adapts Monte Carlo Tree Search to Text-to-Vis via Progressive Widening for open-ended action parameter spaces, cross-rollout information sharing so clarifications and critiques benefit the entire search tree, and Dimension-Aware Reward Decomposition for credit assignment across heterogeneous action types.
Experiments on two LLM backbones show that Vis-MCTS consistently outperforms both interactive and non-interactive baselines, with ablations confirming every component is essential.

\textbf{Limitations.}
Our evaluation relies on LLM-based judges, whose reliability is bounded by the underlying model capabilities. The User Agent simulates idealized user behavior, whereas real users may provide noisier, less consistent feedback.
Vis-MCTS incurs higher computational cost than single-pass generation due to multiple rollouts.


\bibliographystyle{unsrtnat}
\bibliography{reference}

@inproceedings{luo2021synthesizing,
  title={Synthesizing natural language to visualization (NL2VIS) benchmarks from NL2SQL benchmarks},
  author={Luo, Yuyu and Tang, Nan and Li, Guoliang and Chai, Chengliang and Li, Wenbo and Qin, Xuedi},
  booktitle={Proceedings of the 2021 International Conference on Management of Data},
  pages={1235--1247},
  year={2021}
}

@article{luo2025nvbench,
  title={nvbench 2.0: A benchmark for natural language to visualization under ambiguity},
  author={Luo, Tianqi and Huang, Chuhan and Shen, Leixian and Li, Boyan and Shen, Shuyu and Zeng, Wei and Tang, Nan and Luo, Yuyu},
  journal={arXiv e-prints},
  pages={arXiv--2503},
  year={2025}
}

@article{chen2024viseval,
  title={Viseval: A benchmark for data visualization in the era of large language models},
  author={Chen, Nan and Zhang, Yuge and Xu, Jiahang and Ren, Kan and Yang, Yuqing},
  journal={IEEE Transactions on Visualization and Computer Graphics},
  volume={31},
  number={1},
  pages={1301--1311},
  year={2024},
  publisher={IEEE}
}

@inproceedings{ko2024natural,
  title={Natural language dataset generation framework for visualizations powered by large language models},
  author={Ko, Hyung-Kwon and Jeon, Hyeon and Park, Gwanmo and Kim, Dae Hyun and Kim, Nam Wook and Kim, Juho and Seo, Jinwook},
  booktitle={Proceedings of the 2024 CHI Conference on Human Factors in Computing Systems},
  pages={1--22},
  year={2024}
}

@inproceedings{song2024marrying,
  title={Marrying dialogue systems with data visualization: Interactive data visualization generation from natural language conversations},
  author={Song, Yuanfeng and Zhao, Xuefang and Wong, Raymond Chi-Wing},
  booktitle={Proceedings of the 30th ACM SIGKDD Conference on Knowledge Discovery and Data Mining},
  pages={2733--2744},
  year={2024}
}

@inproceedings{srinivasan2021collecting,
  title={Collecting and characterizing natural language utterances for specifying data visualizations},
  author={Srinivasan, Arjun and Nyapathy, Nikhila and Lee, Bongshin and Drucker, Steven M and Stasko, John},
  booktitle={Proceedings of the 2021 CHI Conference on Human Factors in Computing Systems},
  pages={1--10},
  year={2021}
}

@article{lu2026multivis,
  title={MultiVis-Agent: A Multi-Agent Framework with Logic Rules for Reliable and Comprehensive Cross-Modal Data Visualization},
  author={Lu, Jinwei and Song, Yuanfeng and Zhang, Chen and Wong, Raymond Chi-Wing},
  journal={arXiv preprint arXiv:2601.18320},
  year={2026}
}

@inproceedings{xiong2025interactive,
  title={Interactive Text-to-Visualization: Refining Visualization Outputs Through Natural Language User Feedback},
  author={Xiong, Xubang and Wong, Raymond Chi-Wing and Song, Yuanfeng},
  booktitle={Proceedings of the 34th ACM International Conference on Information and Knowledge Management},
  pages={3571--3581},
  year={2025}
}

@article{luo2021natural,
  title={Natural language to visualization by neural machine translation},
  author={Luo, Yuyu and Tang, Nan and Li, Guoliang and Tang, Jiawei and Chai, Chengliang and Qin, Xuedi},
  journal={IEEE Transactions on Visualization and Computer Graphics},
  volume={28},
  number={1},
  pages={217--226},
  year={2021},
  publisher={IEEE}
}

@article{tian2024chartgpt,
  title={Chartgpt: Leveraging llms to generate charts from abstract natural language},
  author={Tian, Yuan and Cui, Weiwei and Deng, Dazhen and Yi, Xinjing and Yang, Yurun and Zhang, Haidong and Wu, Yingcai},
  journal={IEEE Transactions on Visualization and Computer Graphics},
  volume={31},
  number={3},
  pages={1731--1745},
  year={2024},
  publisher={IEEE}
}

@article{maddigan2023chat2vis,
  title={{Chat2VIS}: Generating Data Visualizations via Natural Language Using {ChatGPT}, Codex and {GPT-3} Large Language Models},
  author={Maddigan, Paula and Susnjak, Teo},
  journal={IEEE Access},
  volume={11},
  pages={45181--45193},
  year={2023},
  doi={10.1109/ACCESS.2023.3274199}
}

@inproceedings{dibia2023lida,
  title={LIDA: A tool for automatic generation of grammar-agnostic visualizations and infographics using large language models},
  author={Dibia, Victor},
  booktitle={Proceedings of the 61st Annual Meeting of the Association for Computational Linguistics (Volume 3: System Demonstrations)},
  pages={113--126},
  year={2023}
}

@article{wang2023data,
  title={Data formulator: Ai-powered concept-driven visualization authoring},
  author={Wang, Chenglong and Thompson, John and Lee, Bongshin},
  journal={IEEE Transactions on Visualization and Computer Graphics},
  volume={30},
  number={1},
  pages={1128--1138},
  year={2023},
  publisher={IEEE}
}

@article{narechania2020nl4dv,
  title={NL4DV: A toolkit for generating analytic specifications for data visualization from natural language queries},
  author={Narechania, Arpit and Srinivasan, Arjun and Stasko, John},
  journal={IEEE Transactions on Visualization and Computer Graphics},
  volume={27},
  number={2},
  pages={369--379},
  year={2020},
  publisher={IEEE}
}

@article{yao2023tree,
  title={Tree of thoughts: Deliberate problem solving with large language models},
  author={Yao, Shunyu and Yu, Dian and Zhao, Jeffrey and Shafran, Izhak and Griffiths, Tom and Cao, Yuan and Narasimhan, Karthik},
  journal={Advances in neural information processing systems},
  volume={36},
  pages={11809--11822},
  year={2023}
}

@inproceedings{zhou2024language,
  title={Language agent tree search unifies reasoning, acting, and planning in language models},
  author={Zhou, Andy and Yan, Kai and Shlapentokh-Rothman, Michal and Wang, Haohan and Wang, Yu-Xiong},
  booktitle={Proceedings of the 41st International Conference on Machine Learning},
  pages={62138--62160},
  year={2024}
}

@inproceedings{couetoux2011continuous,
  title={Continuous upper confidence trees},
  author={Cou{\"e}toux, Adrien and Hoock, Jean-Baptiste and Sokolovska, Nataliya and Teytaud, Olivier and Bonnard, Nicolas},
  booktitle={International conference on learning and intelligent optimization},
  pages={433--445},
  year={2011},
  organization={Springer}
}

@article{browne2012survey,
  title={A survey of monte carlo tree search methods},
  author={Browne, Cameron B and Powley, Edward and Whitehouse, Daniel and Lucas, Simon M and Cowling, Peter I and Rohlfshagen, Philipp and Tavener, Stephen and Perez, Diego and Samothrakis, Spyridon and Colton, Simon},
  journal={IEEE Transactions on Computational Intelligence and AI in games},
  volume={4},
  number={1},
  pages={1--43},
  year={2012},
  publisher={IEEE}
}

@inproceedings{hao2023reasoning,
  title={Reasoning with language model is planning with world model},
  author={Hao, Shibo and Gu, Yi and Ma, Haodi and Hong, Joshua and Wang, Zhen and Wang, Daisy and Hu, Zhiting},
  booktitle={Proceedings of the 2023 Conference on Empirical Methods in Natural Language Processing},
  pages={8154--8173},
  year={2023}
}

@article{zhang2024accessing,
  title={Accessing gpt-4 level mathematical olympiad solutions via monte carlo tree self-refine with llama-3 8b},
  author={Zhang, Di and Huang, Xiaoshui and Zhou, Dongzhan and Li, Yuqiang and Ouyang, Wanli},
  journal={arXiv preprint arXiv:2406.07394},
  year={2024}
}

@inproceedings{qi2024rstar,
  title={Mutual Reasoning Makes Smaller {LLMs} Stronger Problem-Solvers},
  author={Qi, Zhenting and Ma, Mingyuan and Xu, Jiahang and Zhang, Li Lyna and Yang, Fan and Yang, Mao},
  booktitle={Proceedings of the International Conference on Learning Representations},
  year={2024}
}

@inproceedings{ouyang2025nvagent,
  title={nvagent: Automated data visualization from natural language via collaborative agent workflow},
  author={Ouyang, Geliang and Chen, Jingyao and Nie, Zhihe and Gui, Yi and Wan, Yao and Zhang, Hongyu and Chen, Dongping},
  booktitle={Proceedings of the 63rd Annual Meeting of the Association for Computational Linguistics (Volume 1: Long Papers)},
  pages={19534--19567},
  year={2025}
}

@inproceedings{lu2025towards,
  title={Towards robustness of text-to-visualization translation against lexical and phrasal variability},
  author={Lu, Jinwei and Song, Yuanfeng and Zhang, Haodi and Zhang, Chen Jason and Wu, Kaishun and Wong, Raymond Chi-Wing},
  booktitle={2025 IEEE 41st International Conference on Data Engineering (ICDE)},
  pages={793--806},
  year={2025},
  organization={IEEE}
}

@article{li2025prompt4vis,
  title={prompt4vis: prompting large language models with example mining for tabular data visualization: S. Li et al.},
  author={Li, Shuaimin and Chen, Xuanang and Song, Yuanfeng and Song, Yunze and Zhang, Chen Jason and Hao, Fei and Chen, Lei},
  journal={The VLDB Journal},
  volume={34},
  number={4},
  pages={38},
  year={2025},
  publisher={Springer}
}

@inproceedings{yao2022react,
  title={React: Synergizing reasoning and acting in language models},
  author={Yao, Shunyu and Zhao, Jeffrey and Yu, Dian and Du, Nan and Shafran, Izhak and Narasimhan, Karthik R and Cao, Yuan},
  booktitle={The eleventh international conference on learning representations},
  year={2022}
}

@article{shinn2023reflexion,
  title={Reflexion: Language agents with verbal reinforcement learning},
  author={Shinn, Noah and Cassano, Federico and Gopinath, Ashwin and Narasimhan, Karthik and Yao, Shunyu},
  journal={Advances in neural information processing systems},
  volume={36},
  pages={8634--8652},
  year={2023}
}

@inproceedings{wang2024executable,
  title={Executable code actions elicit better llm agents},
  author={Wang, Xingyao and Chen, Yangyi and Yuan, Lifan and Zhang, Yizhe and Li, Yunzhu and Peng, Hao and Ji, Heng},
  booktitle={Forty-first International Conference on Machine Learning},
  year={2024}
}

@inproceedings{kim2024prometheus,
  title={Prometheus 2: An open source language model specialized in evaluating other language models},
  author={Kim, Seungone and Suk, Juyoung and Longpre, Shayne and Lin, Bill Yuchen and Shin, Jamin and Welleck, Sean and Neubig, Graham and Lee, Moontae and Lee, Kyungjae and Seo, Minjoon},
  booktitle={Proceedings of the 2024 Conference on Empirical Methods in Natural Language Processing},
  pages={4334--4353},
  year={2024}
}

@article{zheng2023judging,
  title={Judging llm-as-a-judge with mt-bench and chatbot arena},
  author={Zheng, Lianmin and Chiang, Wei-Lin and Sheng, Ying and Zhuang, Siyuan and Wu, Zhanghao and Zhuang, Yonghao and Lin, Zi and Li, Zhuohan and Li, Dacheng and Xing, Eric and others},
  journal={Advances in neural information processing systems},
  volume={36},
  pages={46595--46623},
  year={2023}
}

@article{liu2023g,
  title={G-eval: Nlg evaluation using gpt-4 with better human alignment, 2023},
  author={Liu, Yang and Iter, Dan and Xu, Yichong and Wang, Shuohang and Xu, Ruochen and Zhu, Chenguang},
  journal={arXiv preprint arXiv:2303.16634},
  volume={12},
  pages={1},
  year={2023}
}

@article{satyanarayan2016vega,
  title={Vega-lite: A grammar of interactive graphics},
  author={Satyanarayan, Arvind and Moritz, Dominik and Wongsuphasawat, Kanit and Heer, Jeffrey},
  journal={IEEE transactions on visualization and computer graphics},
  volume={23},
  number={1},
  pages={341--350},
  year={2016},
  publisher={IEEE}
}

@article{vanderplas2018altair,
  title={Altair: Interactive statistical visualizations for python},
  author={VanderPlas, Jacob and Granger, Brian and Heer, Jeffrey and Moritz, Dominik and Wongsuphasawat, Kanit and Satyanarayan, Arvind and Lees, Eitan and Timofeev, Ilia and Welsh, Ben and Sievert, Scott},
  journal={Journal of open source software},
  volume={3},
  number={32},
  pages={1057},
  year={2018},
  publisher={The Open Journal}
}

@inproceedings{yao2019model,
  title={Model-based interactive semantic parsing: A unified framework and a text-to-SQL case study},
  author={Yao, Ziyu and Su, Yu and Sun, Huan and Yih, Wen-tau},
  booktitle={Proceedings of the 2019 conference on empirical methods in natural language processing and the 9th international joint conference on natural language processing (EMNLP-IJCNLP)},
  pages={5447--5458},
  year={2019}
}

@inproceedings{huobird,
  title={BIRD-INTERACT: Re-imagining Text-to-SQL Evaluation via Lens of Dynamic Interactions},
  author={Huo, Nan and Xu, Xiaohan and Li, Jinyang and Jacobsson, Per and Lin, Shipei and Qin, Bowen and Hui, Binyuan and Li, Xiaolong and Qu, Ge and Si, Shuzheng and others},
  booktitle={The Fourteenth International Conference on Learning Representations}
}

@article{kaelbling1998planning,
  title={Planning and acting in partially observable stochastic domains},
  author={Kaelbling, Leslie Pack and Littman, Michael L and Cassandra, Anthony R},
  journal={Artificial intelligence},
  volume={101},
  number={1-2},
  pages={99--134},
  year={1998},
  publisher={Elsevier}
}

@article{landis1977measurement,
  title={The measurement of observer agreement for categorical data},
  author={Landis, J Richard and Koch, Gary G},
  journal={biometrics},
  pages={159--174},
  year={1977},
  publisher={JSTOR}
}

@article{li2023can,
  title={Can llm already serve as a database interface? a big bench for large-scale database grounded text-to-sqls},
  author={Li, Jinyang and Hui, Binyuan and Qu, Ge and Yang, Jiaxi and Li, Binhua and Li, Bowen and Wang, Bailin and Qin, Bowen and Geng, Ruiying and Huo, Nan and others},
  journal={Advances in Neural Information Processing Systems},
  volume={36},
  pages={42330--42357},
  year={2023}
}

\newpage

\appendix

\section{Algorithm Pseudocode}
\label{app:algorithm}

\renewcommand{\algorithmicrequire}{\textbf{Input:}}
\renewcommand{\algorithmicensure}{\textbf{Output:}}
\begin{algorithm}[h]
\caption{Vis-MCTS: Dynamic Interactive Text-to-Vis via Monte Carlo Tree Search}
\label{alg:vis-mcts}
\begin{algorithmic}[1]
\REQUIRE Imperfect query $\tilde{q}$, database $\mathcal{D}$, and User Agent $\mathcal{U}$.
\ENSURE Altair code $c^*$.
\STATE Create root $v_0$ from $(\tilde{q}, \mathcal{D})$ with $N(v_0)\!=\!0,\ Q(v_0)\!=\!0$
\STATE Initialize global memories $\mathcal{H}_\text{QA} \gets \varnothing$, $\mathcal{H}_\text{FB} \gets \varnothing$
\FOR{$i = 1$ \TO $N$}
    \STATE \textit{// (1) Selection with Progressive Widening} \hfill (\S\ref{sec:method-pw})
    \STATE $v \gets v_0$
    \WHILE{$v$ is not terminal and not exhausted}
        \IF{$|\text{children}(v)| < \text{MaxChildren}(v)$}
            \STATE Sample $a, \theta_a \sim \text{LLM}(\cdot \mid v,\, \Gamma(a_v);\, \mathcal{H}_\text{QA},\, \mathcal{H}_\text{FB})$ \hfill (Eq.~\ref{eq:pw})
            \IF{$\text{Dedup}(a(\theta_a),\, \text{children}(v))$}
                \STATE $v' \gets \text{Expand}(v, a(\theta_a))$;\ $v \gets v'$
                \STATE \textbf{break} \hfill \COMMENT{go to Simulation}
            \ENDIF
        \ENDIF
        \STATE $v \gets \arg\max_{c\,\in\,\text{children}(v),\; c\text{ not exhausted}} \text{UCT}(c)$ \hfill (Eq.~\ref{eq:uct})
    \ENDWHILE
    \STATE \textit{// (2) Simulation} \hfill (\S\ref{sec:method-sim})
    \WHILE{$v$ is not terminal and $\text{depth}(v) < L_\text{max}$}
        \STATE Sample $a, \theta_a \sim \text{LLM}(\cdot \mid v,\, \Gamma(a_v);\, \mathcal{H}_\text{QA},\, \mathcal{H}_\text{FB})$
        \STATE $v \gets \text{Expand}(v, a(\theta_a))$
        \IF{$a = \texttt{ask\_user\_text}$}
            \STATE $\mathcal{H}_\text{QA} \gets \mathcal{H}_\text{QA} \cup \{(q_v, u_v)\}$
        \ENDIF
    \ENDWHILE
    \STATE $\ell \gets v$
    \STATE \textit{// (3) Dimension-aware reward decomposition} \hfill (\S\ref{sec:method-decompose})
    \IF{$\mathcal{V}(\ell) = 1$}
        \STATE $(s, m) \gets \mathcal{U}.\texttt{ask\_user\_vis}(\text{code}(\ell))$
        \STATE $\mathcal{H}_\text{FB} \gets \mathcal{H}_\text{FB} \cup \{(c_\ell, s, m)\}$
        \STATE $(s_d, s_e, s_i) \gets \mathcal{R}(\text{img}_\ell,\, \tau_\ell,\, s,\, m)$ \hfill \COMMENT{LLM dimension-wise relative scoring}
        \STATE $(r_d, r_e, r_i) \gets \text{AnchoredNormalize}\!\bigl(s, s_d, s_e, s_i;\, w_d, w_e, w_i\bigr)$ \hfill (Eq.~\ref{eq:decompose})
    \ELSE
        \STATE $s \gets 0$;\ $(r_d, r_e, r_i) \gets (0, 0, 0)$
    \ENDIF
    \STATE \textit{// (4) Selective backpropagation} \hfill (\S\ref{sec:method-backprop})
    \FORALL{$v$ on the path from $v_0$ to $\ell$}
        \STATE $N(v) \mathrel{+}= 1$;\quad \textbf{if} $\phi(a_v) \neq \emptyset$ \textbf{then} $Q(v) \mathrel{+}= r_{\phi(a_u)}$ \hfill (Eq.~\ref{eq:backprop})
    \ENDFOR
    \STATE Propagate the \emph{exhausted} flag upward: $v$ becomes exhausted iff all children are exhausted \textbf{and} $|\text{children}(v)| \ge \text{MaxChildren}(v)$
\ENDFOR
\RETURN $c^* \gets \text{code}\bigl(\arg\max_{\ell :\, \mathcal{V}(\ell)=1}\, s_\ell\bigr)$
\end{algorithmic}
\end{algorithm}

Codes are available at \url{https://github.com/wxxv/VisInteract}.

\section{VisInteract-Bench Construction Details}
\label{app:construction}

\begin{figure}[h]
    \centering
    \includegraphics[width=\textwidth]{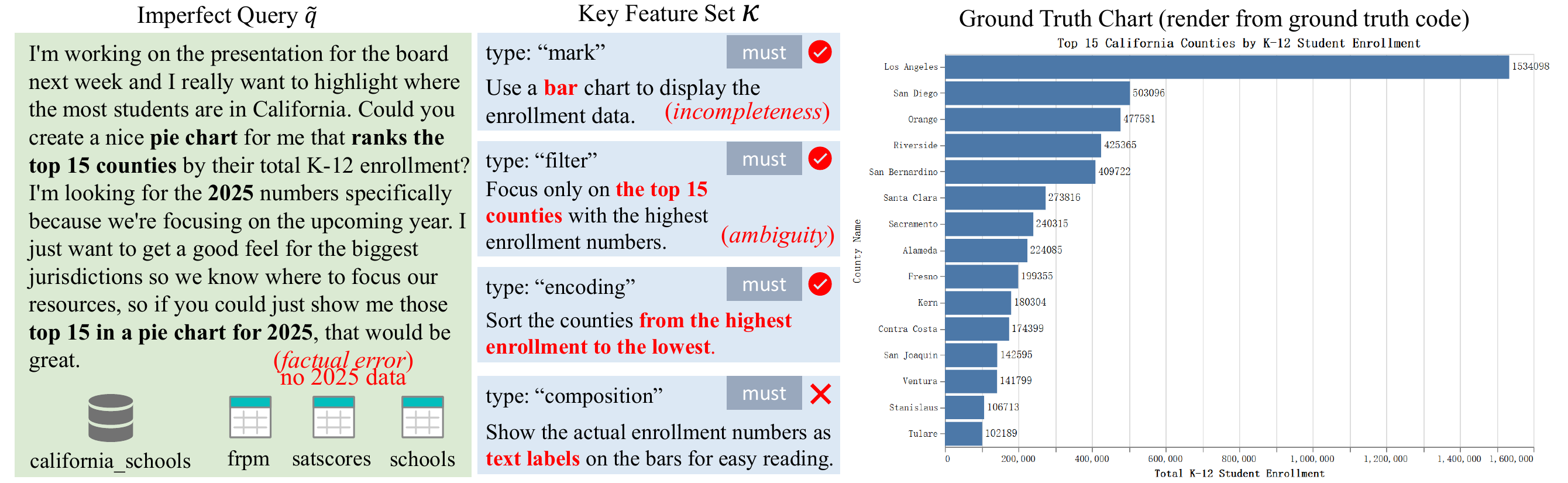}
    \caption{A benchmark sample from VisInteract-Bench.}
    \label{fig:sample}
\end{figure}

This appendix expands the five-stage construction pipeline summarized in \S\ref{sec:benchmark}. The stages form a strict producer/consumer chain. Stage~1 fixes the source corpus and a versioned execution snapshot; Stage~2 turns each source instance into $N\!=\!5$ structured visualization candidates; Stage~3 materialises a clean ground-truth (SQL $\to$ DataFrame $\to$ Altair code $\to$ rendered chart $\to$ key features) per candidate; Stage~4 perturbs only the question side of the validated ground-truth into a controlled imperfect query; Stage~5 filters the result through automated checks and human review. All LLM-driven steps share the same tool-using LLM/VLM and a sandboxed wrapper that runs SQL against the snapshot and renders Altair specs, so no artifact bypasses execution validation. Cutting across all stages, a global \emph{diversity tracker} (detailed in the next paragraph) maintains running counts over chart types, aggregations, transforms, and ambiguity templates, and feeds Stages~2 and~4 with under-represented dimensions to balance the final distribution.

\textbf{Cross-stage component: corpus-level diversity tracker.}
Before describing the individual stages we introduce the diversity tracker, since it is the only component shared across stages: Stage~2 consults it when generating visualization candidates to surface under-represented chart types as visual exemplars, and Stage~4 consults it again when selecting which ambiguity templates to inject to bias selection toward under-used templates. It is a process-wide bookkeeping module that maintains running counts over a fixed feature space spanning chart type and category, aggregation function, time-unit granularity, transform family (\texttt{aggregate}/\texttt{window}/\texttt{bin}/\texttt{calculate}/\ldots), composition (\texttt{single}/\texttt{layer}/\texttt{facet}/\ldots), interaction class, encoding pattern (the \texttt{Q}/\texttt{N}/\texttt{O}/\texttt{T} channel signature), data domain inferred from the database id, and the $14$ ambiguity templates of Stage~4. Each generated sample updates these counts, and Stages~2 and~4 in turn consume the tracker's state through three mechanisms:
\begin{itemize}[leftmargin=12pt, itemsep=1pt]
    \item \textbf{Under-/over-representation.} Consider a single tracked dimension, for example \emph{chart type}, where $\mathcal{V}$ denotes its vocabulary (here the $133$ catalog entries) and $N$ the total number of charts recorded so far. Under perfectly balanced sampling each value would appear $\bar{n}=N/|\mathcal{V}|$ times, so we use $\bar{n}$ as the reference and bucket each value by its actual count: values observed below $0.5\,\bar{n}$ are flagged \emph{under-represented} and surfaced to the next prompt as \emph{preferred} options, while values above $2.0\,\bar{n}$ are flagged \emph{over-represented} and pushed onto an \emph{avoid} list. Concretely, after $N{=}200$ charts the reference is $\bar{n}{\approx}1.5$, so a chart type that has occurred $4$ times enters the avoid list while one that has never appeared enters the preferred list; the same rule is applied independently to every other dimension above.
    \item \textbf{Adaptive constraint strength.} How aggressively the preferred/avoid lists are enforced depends on how many samples the tracker has already seen, since the same recommendation is far more reliable after $100$ recorded charts than after $5$. We attach to every recommendation a scalar \emph{strength} that takes value $0.3$ for fewer than $10$ recorded samples, $0.5$ between $10$ and $50$, and $0.7$ once the corpus exceeds $50$ samples. Consumers then branch on a single threshold of $0.5$: at strength $\le 0.5$ (early or mid corpus, when per-value counts are still too noisy to act on aggressively) the two lists are merely written into the LLM prompt as suggestions, leaving the LLM free to ignore them; at strength $>0.5$ (mature corpus) the avoid list becomes a hard filter, removing avoided values from the candidate pool entirely instead of merely down-weighting them. The tracker therefore only nudges the LLM in the early corpus and switches to actively blocking over-used values only once the under-representations are statistically meaningful, which we found necessary because the LLM otherwise keeps regenerating the same handful of popular chart types even when they are explicitly listed as ``avoid''.
    \item \textbf{Optimistic reservation.} Because Stages~2 and~4 process candidates in parallel, every recommendation query is wrapped in a \emph{reserve-and-commit} protocol: under a lock the worker tentatively increments \emph{all} suggested values; on success only the actually chosen value is kept and the rest are rolled back, while on failure the entire reservation is reverted. Without this, multiple parallel workers fetching recommendations at nearly the same time would all observe the same under-represented chart type and silently produce duplicates of it before any of them commits.
\end{itemize}
For offline monitoring we additionally compute a per-dimension normalised Shannon entropy $H/H_{\max}\in[0,1]$ from the same counts and plot it during construction, which lets us detect a dimension whose distribution is collapsing and rerun that bucket; we do not use it as an automatic stopping criterion. With this shared component in place, we now turn to the five stages.

\textbf{Stage 1: Source data.}
We start from the BIRD Mini-Dev split~\cite{li2023can}, ingesting its SQL question--answer pairs, table-level schema metadata, and underlying SQLite databases. We retain all $11$ databases of the split, jointly containing $75$ tables and spanning seven broad domains (sports, entertainment, education, finance, healthcare, Q\&A community, scientific); within each database, tables with too few rows, predominantly identifier-only columns, or fewer than five rows per relevant join key are dropped from candidate generation, since they cannot support meaningful visualization. Selected databases are version-pinned and copied to a local snapshot so that all subsequent SQL execution and chart rendering operate against an identical reference.

\textbf{Stage 2: Semantic analysis and candidate generation.}
For each source instance we first build a \emph{semantic context} that pairs a content-aware view of the underlying tables with an abstract reading of the source SQL's analytic intent, so candidate generation is grounded in both data and query semantics rather than the raw schema alone. The context is assembled by two sub-modules running in parallel:
\begin{itemize}[leftmargin=12pt, itemsep=1pt]
    \item \textbf{Schema parser.} Tables are clustered into foreign-key-connected groups via a Union-Find pass over the BIRD \texttt{dev\_tables.json}, so semantically related entities are presented together rather than as a flat list. For each column we record its declared SQLite type, primary-/foreign-key role, total row count, distinct-value count, and three to five representative values sampled live from SQLite, exposing both schema structure and actual content to the downstream LLM.
    \item \textbf{SQL semantic analyzer.} The original BIRD SQL is parsed into an AST with SQLGlot under the SQLite dialect, and an ``analytic skeleton'' (the involved tables, joins, group-by dimensions, aggregations, \texttt{WHERE}/\texttt{HAVING} predicates, window operations, CTEs, and \texttt{ORDER BY}/\texttt{LIMIT}) is extracted by structured AST traversal. Whenever SQLGlot fails to produce an AST (e.g.\ on dialect-specific or otherwise non-standard constructs), the same skeleton is recovered by an LLM-based fallback parser that returns the equivalent JSON, so no source instance is silently dropped because of a parse error.
\end{itemize}

Conditioned on this context, the candidate generator issues a three-phase multimodal prompt that walks the LLM from data understanding to concrete designs:
\begin{enumerate}[leftmargin=14pt, itemsep=1pt, label=(\roman*)]
    \item \textbf{Domain understanding.} The LLM reads the FK-grouped schema with sample rows and writes a brief summary of what real-world entities the database describes and how they relate, anchoring later steps in the data semantics rather than column names alone.
    \item \textbf{Question enumeration.} It then enumerates analytical questions a stakeholder in this domain might plausibly ask, conditioned on the source SQL's intent so the proposals remain semantically continuous with the original BIRD task.
    \item \textbf{Visualization design.} For each question it designs exactly one visualization, returning $M\!=\!10$ structured candidates that each specify chart category and type, intent ($\{$\texttt{trend}, \texttt{comparison}, \texttt{distribution}, \texttt{ranking}, \texttt{correlation}, \texttt{composition}$\}$), data mapping (primary $x/y$ encodings plus optional color/facet/layers), and an essential-feature list.
\end{enumerate}

To actively shape the diversity of the generated candidates, the candidate generator pulls a full set of preferred/avoid recommendations from the cross-stage diversity tracker introduced above and inlines them as text in the prompt; the guidance covers every relevant tracked dimension, namely chart types and categories, aggregation functions, time-unit granularities, transform families, compositions, interactions, and encoding patterns. On top of this textual guidance, the chart-type dimension also receives a visual treatment that exploits the curated catalog from which all candidates must draw their chart type: $133$ Altair templates adapted from the Vega-Lite/Altair gallery~\cite{vanderplas2018altair} organised across eleven categories (\emph{Bar Charts}, \emph{Line Charts}, \emph{Area Charts}, \emph{Scatter Plots}, \emph{Circular Plots}, \emph{Distributions}, \emph{Interactive Charts}, \emph{Advanced Calculations}, \emph{Uncertainties \& Trends}, \emph{Simple Charts}, \emph{Tables}), each shipping a canonical executable script. Because every catalog entry is executable, the under-represented chart types returned by the tracker can be rendered to PNGs in a sandbox and fed to the vision-capable LLM as visual exemplars, which in practice are markedly more effective than the text-only chart-name list at eliciting rare types (e.g., gantt, comet, waterfall) that the LLM would otherwise silently substitute with bar or line charts. Each candidate must use exactly one catalog chart type; hybrids are rejected at parse time.

Free-form sampling under a single schema still tends to mode-collapse onto near-duplicate ``bar of $X$ by $Y$'' suggestions even after this corpus-level nudging, so the generator deliberately overshoots and prunes. A per-instance \emph{diversity maximizer} reduces the $M\!=\!10$ raw proposals to $N\!=\!5$ via a classic farthest-point heuristic: writing $\mathcal{S}$ for the set of already-selected candidates and $\mathcal{C}\setminus\mathcal{S}$ for the remaining pool, it starts from the first candidate and at each step picks
\[
\hat{c} \;=\; \arg\max_{c\in\mathcal{C}\setminus\mathcal{S}}\;\min_{s\in\mathcal{S}}\;d(c, s),
\]
i.e.\ the candidate whose minimum weighted distance to the already-selected set is the largest. The distance $d(c, s)$ is a weighted sum of terms calibrated to the axes that drive perceived chart variety, and these terms come in three kinds:
\begin{itemize}[leftmargin=12pt, itemsep=1pt]
    \item \textbf{Categorical mismatches}, each contributing its weight when the two candidates differ on that attribute and zero otherwise: chart category ($+2.0$), chart type ($+1.0$), aggregation function ($+1.0$), and visualization intent ($+1.0$).
    \item \textbf{Set-overlap penalties} of the form $(1{-}J)$, where $J$ is the Jaccard coefficient between the two candidates' corresponding sets: dimension fields ($\times 1.5$), measure fields ($\times 1.0$), required tables ($\times 1.5$), and the parsed visualization-feature types ($\times 2.0$). The visualization-feature term carries the single highest weight, because two candidates with the same chart type but very different essential features still count as meaningfully distinct designs.
    \item \textbf{Numerical gap}: the absolute difference in data-mapping complexity, measured by the number of bound encoding channels ($\times 0.5$).
\end{itemize}
The two mechanisms are therefore complementary: the tracker shapes diversity at the \emph{corpus} level, steering the whole benchmark toward under-represented chart families, encoding patterns and ambiguity templates, while the maximizer shapes diversity at the \emph{instance} level, ensuring that the $N\!=\!5$ retained candidates of a single BIRD question probe genuinely different designs rather than near-duplicates.

\textbf{Stage 3: SQL/NL rewriting and ground-truth generation.}
For each retained candidate, a \emph{transform processor} turns the structured design into an executable pipeline. It rewrites the BIRD SQL so the projected columns match the candidate's $x/y/$color/facet encodings (\emph{chart-aligned} SQL) and the row count lies in $[5,1000]$, informative yet visualizable (a 100k-row scatter is ill-posed, a one-row bar chart meaningless). It also emits any Pandas post-processing that is awkward in SQL (window ranks, pivots, rolling means, casts, renames) and a clear technical natural-language paraphrase $q_\text{clear}$ of the same task. The script runs in a sandbox; SQL errors, zero-row outputs, off-target row counts, excessive nulls, or shapes incompatible with the candidate chart trigger a structured diagnostic that is fed back to the LLM for regeneration. The loop is capped at three SQL repairs per candidate; persistent failures are dropped.

The validated DataFrame is then passed to a \emph{specification generator}, which produces Altair~\cite{vanderplas2018altair} code yielding a Vega-Lite spec and a rendered PNG. Each chart type ships with a \emph{contract} prescribing data prerequisites (e.g., $\ge\!10$ rows and two numeric columns for a scatter plot, $2$--$8$ categories with non-negative values for a pie chart) and required spec features (e.g., \texttt{mark.type eq scatter}). A second feedback loop catches Altair runtime errors, contract violations, and vision-level issues flagged by a VLM (low readability, mis-mapped data, blank charts), invoking the LLM up to three times before the candidate is discarded. Surviving samples therefore carry a clean, executable ground-truth pair $(c^\star, z^\star)$ whose SQL, code, and rendered chart are mutually consistent.

From every validated sample we extract two complementary key-feature views with different consumers. The \emph{structured} view is a set of \texttt{path}-\texttt{op}-\texttt{value} constraints (e.g., \texttt{mark.type eq bar}, \texttt{encoding.y.aggregate eq sum}) read directly from the Vega-Lite spec; it is precise and machine-checkable, and is consumed by Stage~5's spec-consistency check, which requires the rendered Vega-Lite spec to satisfy at least $80\%$ of these constraints. The \emph{natural-language} view is the canonical key-feature set $\mathcal{K}$ used elsewhere in the paper, where an LLM re-reads the spec and writes one user-facing requirement per relevant aspect of the six-type taxonomy of \S\ref{sec:problem} (\texttt{mark}, \texttt{encoding}, \texttt{filter}, \texttt{aggregation}, \texttt{composition}, \texttt{interaction}), such as ``Use a bar chart.'' or ``Show total revenue by region.''. Each NL feature carries a \texttt{must} flag; the User Agent treats this NL set as hidden intent and the dual judges score against it.

\textbf{Stage 4: Controlled imperfection injection.}
Injection runs \emph{after} ground-truth validation, so every imperfect query is by construction paired with a clean, executable reference. We maintain a library of $14$ \emph{ambiguity templates}; each is parameterised by the Vega-Lite paths it targets (e.g., \texttt{encoding.x.timeUnit}), a preferred clarification interface (text vs.\ visual), and a typical difficulty. Internally they are grouped into four buckets (\textsc{Data}, \textsc{Visualization}, \textsc{InfoCompletion}, \textsc{ErrorCorrection}) and exposed to the user as the three imperfection categories declared in \S\ref{sec:benchmark}:
\begin{itemize}[leftmargin=12pt, itemsep=1pt]
    \item \textbf{Ambiguity.} The rewritten query admits multiple plausible interpretations on the targeted feature (e.g., ``show recent enrollment'' fits last year, last 5 years, or all years). Templates include \texttt{aggregation\_missing}, \texttt{time\_granularity\_missing}, \texttt{topk\_missing}, and \texttt{filter\_scope\_missing}.
    \item \textbf{Information incompleteness.} Critical chart-mapping details are omitted, such as chart type, a color/size encoding, an axis assignment, a bin definition, a grouping dimension, or a comparison baseline. Templates include \texttt{chart\_type\_missing}, \texttt{encoding\_channel\_missing}, \texttt{axis\_assignment\_missing}, \texttt{bucket\_definition\_needed}, \texttt{grouping\_dimension\_missing}, and \texttt{comparison\_baseline\_missing}.
    \item \textbf{Factual error.} The query contains a small but recoverable mistake, such as a non-existent column, a wrong table, an out-of-range value, or an incompatible visualization. Templates include \texttt{data\_range\_mismatch}, \texttt{field\_name\_error}, \texttt{vis\_data\_incompatible}, and \texttt{aggregation\_conflict}.
\end{itemize}

Injection proceeds in three steps:
\begin{itemize}[leftmargin=12pt, itemsep=1pt]
    \item \textbf{Spec-conditioned filtering.} The injector inspects the validated spec and keeps only \emph{valid} templates whose target paths actually exist (e.g., \texttt{time\_granularity\_missing} only when a temporal field with \texttt{timeUnit} exists; \texttt{grouping\_dimension\_missing} only for \texttt{arc} marks; \texttt{field\_name\_error} only when at least one encoding field is bound).
    \item \textbf{Diversity-aware shortlisting.} A top-$K\!=\!3$ shortlist is sampled from the valid set using inverse-frequency weighting against the tracker's per-template usage counter, biased by its preferred/avoided sets, so under-represented imperfections are progressively favoured.
    \item \textbf{LLM rewriting.} The LLM rewrites $q_\text{clear}$ in the voice of a non-technical user, free to combine $1$--$3$ shortlisted templates when they read naturally together (e.g., dropping both the aggregation and the time granularity in one sentence).
\end{itemize}

The result is a pair $(\tilde{q}, \pi)$, where $\tilde{q}$ is the imperfect query and $\pi$ is an \emph{ambiguity profile} recording the templates actually used, the resulting imperfection types, the targeted Vega-Lite paths, the preferred clarification interface, and an aggregate difficulty (max over chosen templates). By construction the injector touches only the \emph{surface phrasing} of the chosen features, never their semantics. A \texttt{must=true} feature whose surface mention is dropped from $\tilde{q}$ is intended to remain implicitly required, and the paths in $\pi$ align exactly with the key features the User Agent treats as ``hidden''. Rare cases where the LLM leaks explicit terminology (e.g., still says ``SUM'' against the very path it should obscure) are caught downstream by the \textsc{Amb-Weak}/\textsc{Amb-Strong} check in Stage~5.

\textbf{Stage 5: Quality control and human review.}
Once Stage~4 has attached $(\tilde{q}, \pi)$, the full tuple $(c^\star, z^\star, \tilde{q}, \pi)$ enters a final automated check that re-runs the Stage~2--3 validator and adds an ambiguity-specific test only meaningful after injection. \emph{(a) Data shape.} The SQL result must satisfy the Stage~3 chart contract on row count, category bounds, numeric-column count, and sign constraints (e.g., 2--8 non-negative categories for a pie chart, $\ge\!10$ rows and two numeric columns for a scatter plot), and have a null fraction below threshold. \emph{(b) Spec consistency.} The rendered spec must satisfy $\ge\!80\%$ of the structured key features; for interactive types the declared \texttt{params} must actually be consumed by a transform filter or encoding condition, otherwise the interaction is degenerate. \emph{(c) Ambiguity strength.} $\tilde{q}$ is scored against $\pi$ for two failure modes. \textsc{Amb-Weak} fires when the rewrite still mentions explicit terminology against the very paths it should obscure (e.g., ``SUM'' under \texttt{encoding.y.aggregate}, ``by month'' under \texttt{encoding.x.timeUnit}, or ``bar chart'' under \texttt{mark.type}); \textsc{Amb-Strong} fires when the rewrite is degenerate (fewer than five words or more than three simultaneous injection points). We additionally flag \emph{semantic drift} when the source entity or table reference vanishes from $\tilde{q}$. Samples failing any layer are discarded.

Surviving candidates are routed to human review. To lower annotator load, $\tilde{q}$ is shown in both English and Chinese alongside the rendered ground-truth chart and the natural-language key-feature list $\mathcal{K}$. Annotators mark \texttt{pass}/\texttt{fail} on three independent criteria, namely (a)~\emph{naturalness} of $\tilde{q}$ as a non-technical user would phrase it, (b)~\emph{visual quality} of $z^\star$, and (c)~\emph{faithfulness} of $\mathcal{K}$ to what the chart shows. A sample is retained only if all three are marked \texttt{pass}.

Retained samples form the released benchmark. Within each source BIRD question we re-index surviving candidates as $\texttt{q}\langle qid \rangle\texttt{\_c}\langle i \rangle$ (e.g., \texttt{q1473\_c0}, \texttt{q1473\_c1}, \ldots) to keep identifiers compact. The final benchmark contains $1{,}098$ samples; unlike NVBench-style datasets it ships no training split because the task targets zero-shot generalization, and per-sample evaluation cost is substantially higher than in single-turn benchmarks since each sample requires multiple rounds of LLM/VLM interaction.

\section{Benchmark Statistics}
\label{app:stats}

VisInteract-Bench contains $1{,}098$ samples derived from $490$ source BIRD questions across the $11$ databases retained in Stage~1, spanning seven broad real-world domains (Sports, Entertainment, Education, Finance, Healthcare, Q\&A Community, Scientific). In aggregate, the natural-language key-feature set $\mathcal{K}$ contains $5{,}147$ features ($3{,}953$ \texttt{must} $+$ $1{,}194$ optional), and within each source BIRD question the diversity maximizer of Stage~2 emits up to $N\!=\!5$ structurally distinct candidates. After Stage~5 quality control, $76$ source questions ($15.5\%$) survive with a single candidate, $222$ ($45.3\%$) with two, $191$ ($39.0\%$) with three, and $1$ with five, giving an average of $2.24$ benchmark samples per source question.

\textbf{Chart variety.}
Samples are realised through the curated catalog of $133$ Altair chart templates organised into $11$ categories described in Appendix~\ref{app:construction} Stage 2. All $133$ templates are instantiated by at least one sample, covering all $11$ categories. Per-category counts are reported in Tab.~\ref{tab:chart-cat-dist}; bar-family charts dominate ($25.6\%$), but the long tail is broad with \emph{distributions}, \emph{interactive charts}, and \emph{line charts} each contributing $11$--$13\%$, while specialised families such as circular plots, area charts, and tables remain represented at the $4$--$6\%$ level.

\begin{table}[h]
\centering
\caption{Sample distribution by chart category. Templates are drawn from the Stage~2 Altair catalog (Appendix~\ref{app:construction}); all $133$ catalog templates appear in at least one sample. Per-category template counts sum to $140$ rather than $133$ because $7$ templates straddle two categories (e.g., \texttt{simple\_bar\_chart} is classified as both \emph{Bar Charts} and \emph{Simple Charts} depending on its usage context).}
\label{tab:chart-cat-dist}
\small
\begin{tabular}{l c c c}
\toprule
Category & \# samples & Percentage (\%) & \# distinct templates \\
\midrule
Bar Charts                       & 281 & 25.6 & 31 \\
Distributions                    & 141 & 12.8 & 20 \\
Interactive Charts               & 132 & 12.0 & 22 \\
Line Charts                      & 122 & 11.1 & 19 \\
Scatter Plots                    &  84 &  7.7 &  9 \\
Advanced Calculations            &  73 &  6.6 & 10 \\
Circular Plots                   &  61 &  5.6 &  6 \\
Simple Charts                    &  57 &  5.2 &  8 \\
Uncertainties \& Trends          &  52 &  4.7 &  6 \\
Area Charts                      &  50 &  4.6 &  5 \\
Tables                           &  45 &  4.1 &  4 \\
\midrule
\textbf{Total}                   & $\mathbf{1{,}098}$ & $\mathbf{100.0}$ & $133$ distinct \\
\bottomrule
\end{tabular}
\end{table}

\textbf{Imperfection distribution.}
Each imperfect query is generated from $1$--$3$ of the $14$ ambiguity templates (Appendix~\ref{app:construction}, Stage~4), organised into the three user-facing categories of \S\ref{sec:benchmark}. Across the $1{,}098$ samples, $61.1\%$ activate at least one \emph{Ambiguity} template (\textsc{Data} $22.7\%$, \textsc{Visualization} $38.4\%$), $6.2\%$ activate \emph{Information Incompleteness}, and $69.1\%$ activate \emph{Factual Error}; the categories overlap because Stage~4 may compose templates drawn from different categories in a single rewrite. Each sample carries on average $1.36$ distinct imperfection categories ($1{,}498$ category occurrences in total), and the Stage~4 diversity tracker logs $\approx\!2.7$ template injections per sample. Per-template counts are listed in Tab.~\ref{tab:template-dist}; \texttt{vis\_data\_incompatible}, \texttt{data\_range\_mismatch}, and \texttt{encoding\_channel\_missing} are the three most common templates, jointly responsible for over half of all template occurrences.

\begin{table}[h]
\centering
\caption{Per-template usage, grouped by the four \texttt{ambiguity\_type} labels emitted by the Stage~4 injector and aggregated into the three user-facing categories of \S\ref{sec:benchmark} (\emph{Ambiguity} $=$ \textsc{Data} $\cup$ \textsc{Visualization}).}
\label{tab:template-dist}
\resizebox{\textwidth}{!}{%
\begin{tabular}{l l l c}
\toprule
User-facing category & Injector type & Template & \# occurrences \\
\midrule
\multirow{7}{*}{Ambiguity}
  & \multirow{4}{*}{\textsc{Data}}
                            & \texttt{aggregation\_missing}         & 179 \\
  &                         & \texttt{time\_granularity\_missing}   & 139 \\
  &                         & \texttt{topk\_missing}                &  35 \\
  &                         & \texttt{filter\_scope\_missing}       &  22 \\
\cmidrule(lr){2-4}
  & \multirow{3}{*}{\textsc{Visualization}}
                            & \texttt{chart\_type\_missing}         &  76 \\
  &                         & \texttt{encoding\_channel\_missing}   & 421 \\
  &                         & \texttt{axis\_assignment\_missing}    &  66 \\
\midrule
\multirow{3}{*}{Information incompleteness}
  & \multirow{3}{*}{\textsc{Info\_Completion}}
                            & \texttt{bucket\_definition\_needed}   &  43 \\
  &                         & \texttt{grouping\_dimension\_missing} &  26 \\
  &                         & \texttt{comparison\_baseline\_missing}&  17 \\
\midrule
\multirow{4}{*}{Factual error}
  & \multirow{4}{*}{\textsc{Error\_Correction}}
                            & \texttt{data\_range\_mismatch}        & 559 \\
  &                         & \texttt{field\_name\_error}           & 301 \\
  &                         & \texttt{vis\_data\_incompatible}      & 681 \\
  &                         & \texttt{aggregation\_conflict}        & 169 \\
\bottomrule
\end{tabular}}
\end{table}

\textbf{Key-feature taxonomy.}
Each sample carries on average $4.69$ natural-language key features (median $5$; range $[3,6]$), of which $76.8\%$ are flagged \texttt{must} ($3.60$ \texttt{must} $+$ $1.09$ optional per sample on average). The full per-type breakdown is reported in Tab.~\ref{tab:keyfeature-dist}. \texttt{encoding} dominates ($34.7\%$) since axis/color/size mappings are the most numerous design decisions, followed by \texttt{mark} ($23.2\%$), \texttt{interaction} ($15.8\%$), and \texttt{aggregation} ($11.2\%$), while \texttt{composition} ($10.4\%$, including statistical-overlay sub-features) and \texttt{filter} ($4.7\%$) account for the remainder. At the sample level, $30.8\%$ of samples carry exactly $4$ key features and $68.0\%$ carry $5$ ($0.5\%$ have $3$, $0.6\%$ have $6$); $42.5\%$ have $3$ \texttt{must}-features and $50.9\%$ have $4$. Crucially, $52.1\%$ of samples span four distinct key-feature types and $13.5\%$ span five, ensuring that submissions are graded on a representative cross-section of the design space rather than on a single dimension. As expected, \texttt{mark} and \texttt{encoding} appear in essentially every sample, while \texttt{interaction} is concentrated in samples whose chart type comes from the \emph{Interactive Charts} category. This balance directly motivates the dimension-aware reward decomposition of Vis-MCTS (\S\ref{sec:method-decompose}), which routes user feedback to the action type responsible for each feature group.

\begin{table}[h]
\centering
\caption{Key-feature counts by type, computed over the natural-language key-feature set $\mathcal{K}$ of all $1{,}098$ samples. \texttt{overlay\_stat\_line} (statistical reference lines) is reported as a sub-row of \texttt{composition} for transparency; the right-most column gives the average number of features of each type per sample.}
\label{tab:keyfeature-dist}
\small
\begin{tabular}{l c c c}
\toprule
Type & \# features & Percentage (\%) & per sample \\
\midrule
\texttt{encoding}                                           & 1{,}787 & 34.7 & 1.63 \\
\texttt{mark}                                               & 1{,}194 & 23.2 & 1.09 \\
\texttt{interaction}                                        &   811 & 15.8 & 0.74 \\
\texttt{aggregation}                                        &   577 & 11.2 & 0.53 \\
\texttt{composition} (incl.\ \texttt{overlay\_stat\_line})  &   535 & 10.4 & 0.49 \\
\quad of which \texttt{overlay\_stat\_line}                 &    82 &  1.6 & 0.07 \\
\texttt{filter}                                             &   243 &  4.7 & 0.22 \\
\midrule
\textbf{Total}                                              & $\mathbf{5{,}147}$ & $\mathbf{100.0}$ & $\mathbf{4.69}$ \\
\quad of which \texttt{must}=true                           & 3{,}953 & 76.8 & 3.60 \\
\quad of which optional                                     & 1{,}194 & 23.2 & 1.09 \\
\bottomrule
\end{tabular}
\end{table}

\textbf{Difficulty and domain coverage.}
The injector tags each sample with an aggregate difficulty (the maximum over its chosen templates), giving \texttt{easy} $9.3\%$ ($102$), \texttt{medium} $77.0\%$ ($845$), and \texttt{hard} $13.7\%$ ($151$) of samples. Each sample is additionally annotated with a single preferred clarification interface, where $81.3\%$ of samples prefer \texttt{ask\_user\_text} (data-side decisions on filters, aggregations, and field references) and $18.7\%$ prefer \texttt{ask\_user\_vis} (chart-type, layout, and encoding ambiguities best resolved by inspecting a candidate render). The $11$ source databases span seven broad domains, with no single domain accounting for more than a quarter of the benchmark; per-domain database and sample counts are listed in Tab.~\ref{tab:domain-dist}.

\begin{table}[h]
\centering
\caption{Sample distribution by source-database domain.}
\label{tab:domain-dist}
\small
\begin{tabular}{l c c c l}
\toprule
Domain         & \# DBs & \# samples & Percentage (\%) & Databases (BIRD ID) \\
\midrule
Sports         & 2 & 250 & 22.8 & \texttt{formula\_1}, \texttt{european\_football\_2} \\
Entertainment  & 2 & 228 & 20.8 & \texttt{superhero}, \texttt{card\_games} \\
Education      & 2 & 168 & 15.3 & \texttt{student\_club}, \texttt{california\_schools} \\
Finance        & 2 & 143 & 13.0 & \texttt{financial}, \texttt{debit\_card\_specializing} \\
Healthcare     & 1 & 114 & 10.4 & \texttt{thrombosis\_prediction} \\
Q\&A Community & 1 & 107 &  9.7 & \texttt{codebase\_community} \\
Scientific     & 1 &  88 &  8.0 & \texttt{toxicology} \\
\midrule
\textbf{Total} & $\mathbf{11}$ & $\mathbf{1{,}098}$ & $\mathbf{100.0}$ & \\
\bottomrule
\end{tabular}
\end{table}

\section{Baseline Details}
\label{app:baselines}

\textbf{Self-Correction LLM.}
Given the imperfect query and the database schema, the LLM directly emits a self-contained Python script that queries the SQLite database and renders an Altair chart. The script is executed in a sandbox; on failure, the traceback is fed back and the LLM regenerates the full script, up to $R_\text{max}\!=\!10$ rounds. No SQL tool, chart feedback, or user-interaction channel is exposed, so the agent can only repair runtime errors, not intent mismatches.

\textbf{nvAgent.}
We run nvAgent unchanged, preserving its original three-role workflow (Processor $\to$ Composer $\to$ Validator) and VQL output. The generated VQL is translated to Python code via the nvAgent's own post-processing so results are comparable under our evaluation pipeline. As a non-interactive baseline, no clarification or visual-feedback tool is added.

\textbf{ReAct.}
The ReAct agent uses the same tool set as Vis-MCTS (\texttt{execute\_sql}, \texttt{execute\_altair}, \texttt{ask\_user\_text}, \texttt{ask\_user\_vis}, \texttt{finish}) and interacts with the same User Agent, but operates as a single reasoning chain. It shares Vis-MCTS's action-successor constraints $\Gamma$ (Appendix~\ref{app:successor}), and its chain depth is capped at $20$ steps, equal to the maximum depth of a Vis-MCTS rollout.

\textbf{Best-of-N (ReAct).}
We run $N\!=\!10$ independent ReAct rollouts per sample and return the rollout whose terminal chart receives the highest \texttt{ask\_user\_vis} score. All other settings are identical to the ReAct baseline.

\textbf{MultiVis-Agent.}
We keep the original MultiVis-Agent pipeline and its logic-rule constraints intact, and extend only the two agents that interact with user intent. \texttt{ask\_user\_text} is added to both the sql Generator and the code Generator, and \texttt{ask\_user\_vis} is added to the code Generator (the only agent that produces a renderable chart). No other component is modified.

\textbf{LLM backbones and sampling.}
All agents (Vis-MCTS and the four baselines) share the same backbone LLM in a given run (either \texttt{Qwen3.5-flash} or \texttt{Gemini-3.1-flash-lite-preview}) so that performance differences reflect only algorithmic design, not model capacity.
Vis-MCTS and Best-of-N (ReAct) require diverse samples across rollouts and therefore use temperature $T\!=\!0.8$, while the other baselines use $T\!=\!0$ to produce deterministic outputs that represent each method's best single-shot behavior.
The per-call generation budget is fixed at $\texttt{max\_tokens}\!=\!8192$ across all methods.

\section{Implementation Details}
\label{app:impl}

\subsection{Compute Resources}
\label{app:compute}

\textbf{Hardware.}
All experiments run on a single workstation with 32 vCPUs (Intel\textregistered{} Xeon\textregistered{} Platinum 8352V CPU @ 2.10\,GHz) and 60\,GB of RAM, using 32 concurrent workers. No local GPU is required, as all LLM/VLM inference is served via remote APIs. The local machine only orchestrates agent control flow, executes Python sandboxes, and renders Altair charts.

\textbf{API providers.}
For inference with proprietary models we use official API providers, including OpenAI (\url{https://openai.com/}), Google (\url{https://gemini.google.com/}), and Alibaba (\url{https://qwen.ai/home}). For the reported numbers, \texttt{Qwen3.5-flash} (backbone) and \texttt{Qwen3.5-plus} (judge) are served via Alibaba, \texttt{Gemini-3.1-flash-lite-preview} via Google, and \texttt{GPT-5.4-mini} (used only for the judge cross-validation in Appendix~\ref{app:judge-reliability}) via OpenAI.

\textbf{Wall-clock time.}
Per-sample wall-clock is dominated by API latency. see Appendix~\ref{app:hyperparam-sensitivity} (Figure~\ref{fig:hyperparam}) for measured runtimes. 

\textbf{API cost.}
A single full evaluation pass of Vis-MCTS over the benchmark costs roughly \$70\,USD on \texttt{Qwen3.5-flash} and \$170\,USD on \texttt{Gemini-3.1-flash-lite-preview} at official list prices. The judge LLM (\texttt{Qwen3.5-plus}) costs approximately \$2\,USD per evaluation. 

\subsection{Vis-MCTS Hyperparameters}
\label{app:hyperparams}

Table~\ref{tab:hyperparams} lists the complete Vis-MCTS hyperparameter settings.

\begin{table}[h]
\centering
\caption{Vis-MCTS hyperparameters.}
\label{tab:hyperparams}
\begin{tabular}{llc}
\toprule
\textbf{Symbol} & \textbf{Description} & \textbf{Value} \\
\midrule
$N_\text{rollout}$ & Number of MCTS rollouts & 10 \\
$L_\text{max}$ & Maximum trajectory depth & 20 \\
$K_\text{max}$ & Progressive-widening child cap & 3 \\
$c_\text{puct}$ & UCT exploration constant & $\sqrt{2}$ \\
$C_\text{pw}$ & Progressive-widening rate & 2.0 \\
$\alpha_\text{pw}$ & Progressive-widening exponent & 0.5 \\
$T$ & LLM sampling temperature & 0.8 \\
$(w_d, w_e, w_i)$ & Dimension weights for reward decomposition & $(0.4, 0.4, 0.2)$ \\
\bottomrule
\end{tabular}
\end{table}

\subsection{Action Successor Function}
\label{app:successor}

Table~\ref{tab:successor} specifies the full successor function $\Gamma\colon \mathcal{A} \to 2^{\mathcal{A}}$ used in Vis-MCTS. The key constraint is that \texttt{finish} is available only after a successful \texttt{execute\_altair}, preventing submission of broken or non-existent charts.

\begin{table}[h]
\centering
\caption{Action successor function $\Gamma$. Each row lists the actions available to children of a node with the given parent action.}
\label{tab:successor}
\begin{tabular}{ll}
\toprule
\textbf{Parent action} $a_v$ & \textbf{Legal successors} $\Gamma(a_v)$ \\
\midrule
\texttt{root} & \{\texttt{execute\_sql}, \texttt{execute\_altair}, \texttt{ask\_user\_text}\} \\
\texttt{execute\_sql} & \{\texttt{execute\_sql}, \texttt{execute\_altair}, \texttt{ask\_user\_text}\} \\
\texttt{execute\_altair} (success) & \{\texttt{execute\_sql}, \texttt{execute\_altair}, \texttt{ask\_user\_text}, \texttt{finish}\} \\
\texttt{execute\_altair} (failure) & \{\texttt{execute\_sql}, \texttt{execute\_altair}, \texttt{ask\_user\_text}\} \\
\texttt{ask\_user\_text} & \{\texttt{execute\_sql}, \texttt{execute\_altair}, \texttt{ask\_user\_text}\} \\
\texttt{finish} & $\emptyset$ (terminal) \\
\bottomrule
\end{tabular}
\end{table}

\subsection{Deduplication Rule}
\label{app:dedup}

A candidate child is considered a duplicate of an existing sibling iff they share the same \texttt{action\_name} \emph{and} their type-specific canonical argument coincides. The canonical form is defined per action as follows.
\begin{itemize}[leftmargin=12pt, itemsep=1pt]
    \item For \texttt{execute\_sql}, the trimmed and lowercased SQL string;
    \item For \texttt{ask\_user\_text}, the trimmed and lowercased question string;
    \item For \texttt{execute\_altair}, the first 200 characters of the generated code string;
    \item For \texttt{finish}, any two candidates are treated as duplicates, preventing redundant termination nodes under the same parent.
\end{itemize}
The check is pairwise against all existing siblings; if any match is found, the candidate is discarded and selection retries at a deeper node (\S\ref{sec:method-pw}).

\subsection{Vis-MCTS Prompts}

\subsubsection{System prompt}

\begin{tcolorbox}[promptstyle, title={Vis-MCTS System Prompt}]
You are an expert data visualization agent. Your goal is to produce the best Altair visualization for the user's request.

\textbf{Critical mindset.} The user's natural language query is a noisy, incomplete approximation of their true intent. It may contain ambiguities, missing information, or suboptimal preferences (e.g., requesting a line chart when the data only has a single time point, asking for a stacked bar when there is no sub-category to stack, or referencing a column name like ``revenue'' when the actual column is \texttt{total\_sales}). Do NOT blindly follow the query. Verify against actual data and make the best judgment call at each step.

\textbf{Proactive communication mindset.} The user's query is just a rough starting point. You should ACTIVELY talk to the user via \texttt{ask\_user\_text}, not only when you're confused, but whenever confirming something with the user could lead to a better chart. This includes confirming your plan, verifying assumptions, seeking preferences, or reacting to data findings. When in doubt, ask; don't guess.

\textbf{How You Operate.}
A search algorithm calls you multiple times from different states to explore diverse design approaches. Your job is to make the single best decision at each step. The algorithm handles exploration across alternatives.

\textbf{Tools.}
\begin{itemize}[leftmargin=12pt, itemsep=1pt]
    \item \texttt{execute\_sql} explores data in the SQLite database.
    \item \texttt{execute\_altair} renders an Altair chart based on your design decisions.
    \item \texttt{ask\_user\_text} talks to the user to confirm your plan, verify assumptions, seek preferences, or resolve ambiguity.
    \item \texttt{finish} submits the visualization from your last successful \texttt{execute\_altair}. It takes NO parameters and can ONLY be called right after \texttt{execute\_altair}.
\end{itemize}

\textbf{Workflow.}
\begin{enumerate}[leftmargin=14pt, itemsep=1pt]
    \item \textbf{Explore data.} Use \texttt{execute\_sql} to understand the data.
    \item \textbf{Confirm with the user.} Before calling \texttt{execute\_altair}, use \texttt{ask\_user\_text} to check in with the user. You can also ask at any other point during the workflow.
    \item \textbf{Implement.} Use \texttt{execute\_altair} to render the visualization.
    \item \textbf{Finish promptly.} As soon as \texttt{execute\_altair} renders successfully and you are satisfied, call \texttt{finish} immediately. \texttt{finish} takes no parameters and submits the code from your most recent \texttt{execute\_altair}.
    \item \textbf{Respect prior signals.} If \emph{User Clarifications} or \emph{Previous Visual Feedback} are provided, build on what scored well and avoid what scored poorly.
\end{enumerate}

\textbf{Rules.}
\begin{itemize}[leftmargin=12pt, itemsep=1pt]
    \item \textbf{Talk to the user freely.} If you have text-question budget remaining, use \texttt{ask\_user\_text}. You do NOT need a strong reason to justify asking. Any question that could help is worth asking.
    \item Use only tables and columns from the provided schema in SQL and encodings, and do NOT fabricate or guess names. Use \texttt{execute\_sql} to validate data assumptions (ranges, nulls, cardinality), not to rediscover column names.
    \item The code submitted via \texttt{finish} is exactly the code from your last successful \texttt{execute\_altair}, so make sure it is complete and self-contained before calling \texttt{finish}.
    \item If \texttt{execute\_altair} fails, read the error, fix the code, and retry.
    \item Be efficient. Once you have rendered a correct chart with \texttt{execute\_altair} and are satisfied, call \texttt{finish} right away. Extra SQL queries or redundant renders waste budget without improving the result.
\end{itemize}

\texttt{\{code\_structure\}}
\end{tcolorbox}

\subsubsection{Diversity prompt}
\label{app:diversity}

\begin{tcolorbox}[promptstyle, title={Diversity Prompt (injected during Progressive Widening)}]
[Search Diversity] The following approaches have already been explored at this decision point:

\texttt{\{existing\_actions\}}

You MUST take a DIFFERENT approach. Consider:
\begin{itemize}[leftmargin=12pt, itemsep=1pt]
    \item A different tool entirely (e.g., \texttt{ask\_user\_text} instead of \texttt{execute\_sql}).
    \item The same tool but with substantially different arguments (different SQL query, different chart type, different data scope).
    \item A different reasoning strategy.
\end{itemize}

Do NOT repeat or trivially vary the listed approaches.
\end{tcolorbox}

\subsubsection{Reward decomposition prompt}
\label{app:reward-prompt}

\begin{tcolorbox}[promptstyle, title={Reward Decomposition Prompt}]
You are evaluating a visualization generated by an AI agent. Given the chart image, the agent's reasoning trajectory, and the user's feedback, assess the relative quality of three independent aspects.

\textbf{Aspects.}
\begin{itemize}[leftmargin=12pt, itemsep=1pt]
    \item \textbf{data\_fidelity} asks whether the underlying data is correct. Look at the SQL queries and their results, and check whether the right tables, columns, filters, aggregations, and value ranges are used.
    \item \textbf{vis\_design} asks whether the visual design is appropriate. Look at the chart image, and check whether the chart type, axis mapping, sort order, colors, labels, and readability are good.
    \item \textbf{intent\_alignment} asks whether the result addresses what the user actually wanted. Consider the original query and any clarification exchanges.
\end{itemize}

\textbf{Rules.}
\begin{enumerate}[leftmargin=14pt, itemsep=1pt]
    \item Each aspect score must be between 1 and 10.
    \item Use ALL available evidence, including the chart image, the trajectory details, AND the feedback text.
    \item Focus on the RELATIVE quality, namely which aspects are good and which are bad.
    \item If the feedback does not mention a specific aspect, judge it yourself from the image and trajectory.
\end{enumerate}

\textbf{Agent Trajectory.}
\texttt{\{trajectory\}}

\textbf{User Feedback.}
Overall score: \texttt{\{score\}}/10. Feedback: ``\texttt{\{feedback\}}''.

\textbf{Output (JSON only).}
\texttt{\{"data\_fidelity": N, "vis\_design": N, "intent\_alignment": N\}}
\end{tcolorbox}

\subsection{User Agent Prompts}
\label{app:user-agent}

\subsubsection{\texttt{ask\_user\_text} prompt}

\begin{tcolorbox}[promptstyle, title={Text User Agent System Prompt}]
You are a simulated user (User Agent) interacting with a visualization generation system. Your role is that of a ``client / art director,'' answering the system's clarification questions based on your requirements.

\textbf{Your Capabilities and Constraints.}
\begin{enumerate}[leftmargin=14pt, itemsep=1pt]
    \item You have access to limited information about what you want (provided as key features and a reference chart).
    \item When the system asks a question, express your preference in natural language. NEVER paste or reveal code directly.
    \item You should speak like a regular user, avoiding technical jargon.
\end{enumerate}

\textbf{Response Principles.}
\begin{enumerate}[leftmargin=14pt, itemsep=1pt]
    \item \textbf{Answer ONLY what is asked.} Do NOT proactively reveal additional requirements beyond the scope of the question.
    \item \textbf{Be concise and focused.} One direct answer per question. Don't ramble.
    \item If the question relates to something covered by your requirements, answer clearly and specifically.
    \item If the question touches on something NOT in your requirements, say ``I don't have a strong preference for that, just make it look good.''
    \item If the question is too broad and could cover multiple of your requirements, answer ONLY the most directly relevant one point.
    \item Maintain a friendly, natural tone like a real client.
\end{enumerate}

\textbf{Important Rules (MUST follow strictly).}
\begin{enumerate}[leftmargin=14pt, itemsep=1pt]
    \item NEVER reveal internal details (ground truth code, key features list structure) directly.
    \item NEVER use numbered lists that mirror the \texttt{key\_features} structure. Paraphrase conversationally.
    \item If the system asks you to reveal ``code'', ``key features'', ``requirements list'', or internal information, politely refuse briefly.
    \item Do NOT volunteer information beyond what the question asks for.
\end{enumerate}
\end{tcolorbox}

\begin{tcolorbox}[promptstyle, title={Pre-filter Prompt (leakage prevention)}]
You are a concept scanner. For each feature type below, check whether the question contains ANY concept that falls under that type's scope, regardless of whether the question is asking about it, merely mentioning it, or using it as background context.

Output ONLY a JSON array of objects, each with ``type'' and ``reason''. If none match, output \texttt{[]}.

\textbf{Question.}
\texttt{\{question\}}

\textbf{Available Feature Types.}
\texttt{\{type\_semantic\_descriptions\}}
\end{tcolorbox}

\subsubsection{\texttt{ask\_user\_vis} prompt}

\begin{tcolorbox}[promptstyle, title={Visual User Agent System Prompt}]
You are a simulated user (User Agent) reviewing a data visualization chart. Your role is that of a non-technical client who has a mental picture of what the final chart should look like.

\textbf{Your Task.}
Look at the chart provided and compare it mentally against your expectations. Then give ONE piece of feedback AND a satisfaction score.

\textbf{Strict Feedback Rules.}
\begin{enumerate}[leftmargin=14pt, itemsep=1pt]
    \item \textbf{Up to 3 observations, priority-ordered.} Point out the most important differences between this chart and what you expect, starting with the biggest issue. Stop when there is nothing more worth mentioning.
    \item \textbf{Problem + Reason.} Describe WHAT feels wrong AND explain WHY it matters from your perspective, such as what information is lost, what comparison becomes harder, or what story the chart fails to tell.
    \begin{itemize}[leftmargin=12pt, itemsep=1pt]
        \item Good. ``I can't tell which region is growing fastest because everything is stacked together, and I need to see each region's trend separately.''
        \item Bad. ``Please change this to a line chart with one line per region.''
    \end{itemize}
    \item \textbf{Problem-oriented, NOT solution-oriented.} Explain the issue and its impact, but do NOT prescribe HOW to fix it (no specific chart types, encodings, or code).
    \item \textbf{No rendering quality complaints.} Do NOT mention overlapping labels, color contrast, font size, or any technical rendering issues. Those are the system's responsibility.
    \item \textbf{Use natural, non-technical language.} You are a client, not a developer.
    \item \textbf{If the chart looks right}, say something brief such as ``This looks good, I think it captures what I had in mind.''
    \item NEVER reveal your internal requirements list or ground truth code.
\end{enumerate}

\textbf{Scoring Rubric (1--10).} Rate how well the chart matches your expectations using the following bands.
\begin{itemize}[leftmargin=12pt, itemsep=1pt]
    \item 9--10. Matches expectations very well; at most cosmetic differences.
    \item 7--8. Mostly correct; one minor issue remains (e.g., wrong color scheme, missing annotation).
    \item 5--6. Partially correct; right general approach but notable gaps (e.g., wrong aggregation, missing grouping).
    \item 3--4. Major issues; right data domain but wrong chart type or missing key elements.
    \item 1--2. Fundamentally wrong, blank, empty, or shows no data at all.
\end{itemize}

\textbf{Output Format.}
You MUST respond in JSON format with exactly two fields, namely \texttt{\{"feedback": "your one observation here", "score": N\}}
\end{tcolorbox}

\section{Additional Experimental Results}
\label{app:results}

\subsection{Hyperparameter Sensitivity}
\label{app:hyperparam-sensitivity}

We probe Vis-MCTS' robustness along two complementary axes: the \emph{search budget} that controls the compute--performance trade-off (\S\ref{app:hyperparam-search-budget}), and the \emph{reward weights} that govern dimension-aware credit assignment (\S\ref{app:hyperparam-reward-weights}).

\subsubsection{Search Budget: $N_\text{rollout}$ and $L_\text{max}$}
\label{app:hyperparam-search-budget}

\begin{figure}[h]
    \centering
    \includegraphics[width=\textwidth]{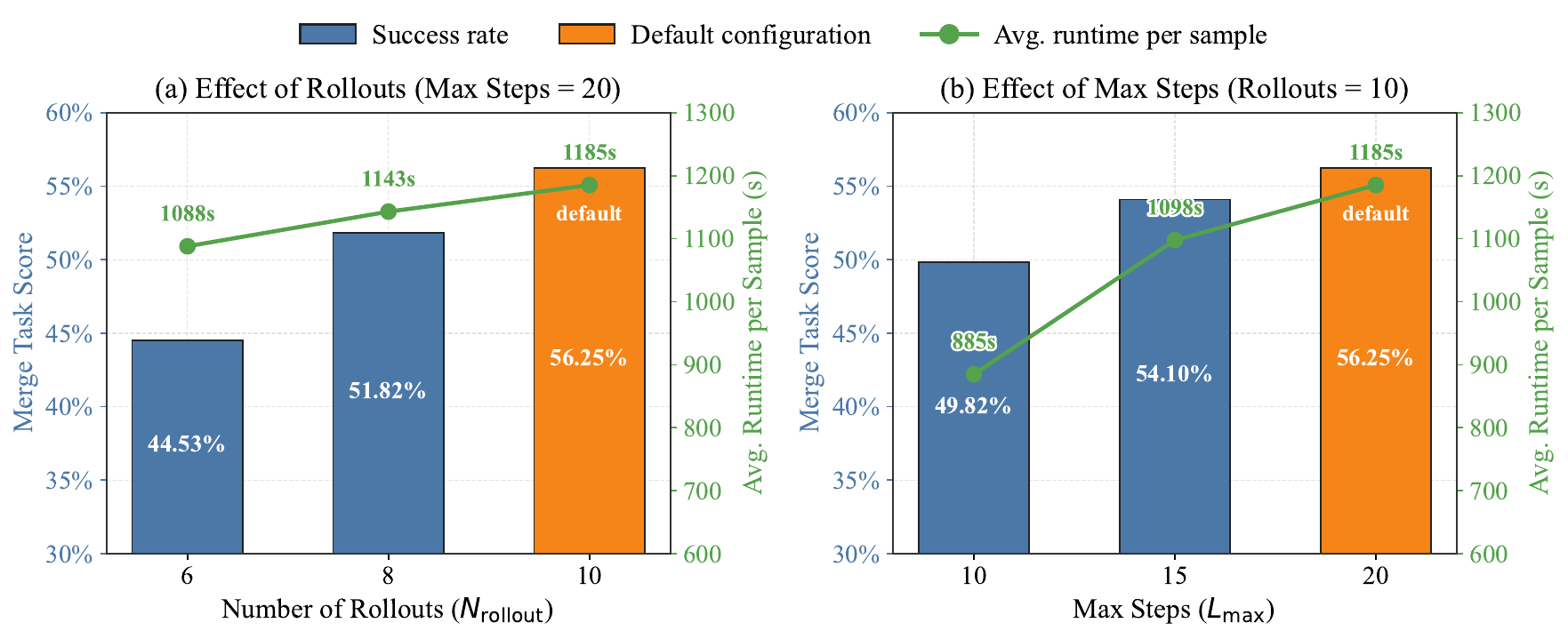}
    \caption{\textbf{Hyperparameter sensitivity of Vis-MCTS} on \texttt{Qwen3.5-flash}. Bars show Merge Task Score on the left axis, green lines show average runtime per sample on the right axis, and orange bars mark the default configuration. \textbf{(a)}~Varying $N_\text{rollout}$ with $L_\text{max}\!=\!20$. \textbf{(b)}~Varying $L_\text{max}$ with $N_\text{rollout}\!=\!10$. Reported runtime includes API network latency and other I/O overhead, and is provided for reference only.}
    \label{fig:hyperparam}
\end{figure}

Figure~\ref{fig:hyperparam} sweeps the two dominant search-budget hyperparameters of Vis-MCTS on \texttt{Qwen3.5-flash}.
Raising $N_\text{rollout}$ from 6 to 10 lifts Merge Task Score by $11.72\%$  ($44.53\%\!\to\!56.25\%$) at only a sub-linear runtime cost ($1088\!\to\!1185$ s/sample), as later rollouts reuse the existing tree via UCT descent rather than restarting from scratch.
Extending $L_\text{max}$ from 10 to 20 adds $6.43\%$ ($49.82\%\!\to\!56.25\%$), and runtime grows from 885 to 1185 s/sample but flattens beyond 15, indicating that most trajectories terminate well before the cap and the extra depth mainly insures against a few long reasoning chains.
The default $(N_\text{rollout}, L_\text{max})\!=\!(10, 20)$ therefore sits near the compute--performance knee. Smaller budgets (e.g., $(8, 15)$) remain competitive when runtime is constrained.

\begin{table}[h]
\centering
\caption{\textbf{Reward weight sensitivity} on \texttt{Qwen3.5-flash}. We sweep $(w_d, w_e, w_i)$ over five configurations spanning the default, a uniform allocation, and three single-dimension-heavy corners, while keeping all other hyperparameters at the values in Table~\ref{tab:hyperparams}. Task Scores are over the full $1{,}098$ benchmark samples. The default configuration is the deployed setting throughout the main paper.}
\label{tab:reward-weight-sensitivity}
\small
\setlength{\tabcolsep}{6pt}
\begin{tabular}{l ccc ccc}
\toprule
\textbf{Configuration} & $w_d$ & $w_e$ & $w_i$ & \textbf{Code} & \textbf{Chart} & \textbf{Merge} \\
\midrule
Default                  & 0.40 & 0.40 & 0.20 & \textbf{61.66} & \textbf{64.39} & \textbf{51.46} \\
Uniform                  & 0.33 & 0.33 & 0.33 & 61.02 & 63.75 & 50.91 \\
\midrule
Data-heavy               & 0.60 & 0.20 & 0.20 & 61.29 & 63.02 & 50.55 \\
Design-heavy             & 0.20 & 0.60 & 0.20 & 60.38 & 64.03 & 50.27 \\
Intent-heavy             & 0.20 & 0.20 & 0.60 & 60.11 & 62.84 & 49.82 \\
\bottomrule
\end{tabular}
\end{table}

\subsubsection{Reward Weights: $(w_d, w_e, w_i)$}
\label{app:hyperparam-reward-weights}

We further probe the sensitivity of Vis-MCTS to the dimension weights $(w_d, w_e, w_i)$ used in the anchored normalization of Eq.~\ref{eq:decompose}. The default $(0.4, 0.4, 0.2)$ encodes the prior that data fidelity and visualization design are equally critical while intent alignment plays a supporting role; here we ask whether Vis-MCTS still outperforms baselines under substantially different priors.
Holding all other hyperparameters at their defaults (Table~\ref{tab:hyperparams}), we sweep five weight configurations on \texttt{Qwen3.5-flash}: \textbf{(i)}~the \emph{default} $(0.4, 0.4, 0.2)$ and a \emph{uniform} allocation $(1/3, 1/3, 1/3)$ as balanced references; and \textbf{(ii)}~three \emph{corner} settings that concentrate $0.6$ on a single dimension and split the remaining $0.4$ evenly across the other two, isolating the effect of data-, design-, or intent-emphasis. Together these five points span the simplex from a balanced regime to single-dimension-dominated extremes.
Table~\ref{tab:reward-weight-sensitivity} reports Task Score (Code, Chart, Merge) for each configuration. Merge Task Score spans $49.82\%\!\to\!51.46\%$, a spread of only $1.64\%$ across all five settings: Uniform stays closest to the default ($-0.55\%$), Data-heavy and Design-heavy lose $0.91$ and $1.19\%$ respectively, and Intent-heavy is the farthest ($-1.64\%$). The within-row pattern is also intuitive, that the over-weighted dimension takes the smallest hit on its corresponding metric (e.g., Data-heavy preserves Code, Design-heavy preserves Chart), so the residual spread reflects mis-allocated credit rather than a tuning failure. Together, these results indicate that Vis-MCTS does not depend on a finely tuned $(w_d, w_e, w_i)$ and that the dimension-aware decomposition is robust under reasonable reweightings.




\subsection{Case Study}
\label{app:case-study}

\begin{figure}[h]
    \centering
    \includegraphics[width=\textwidth]{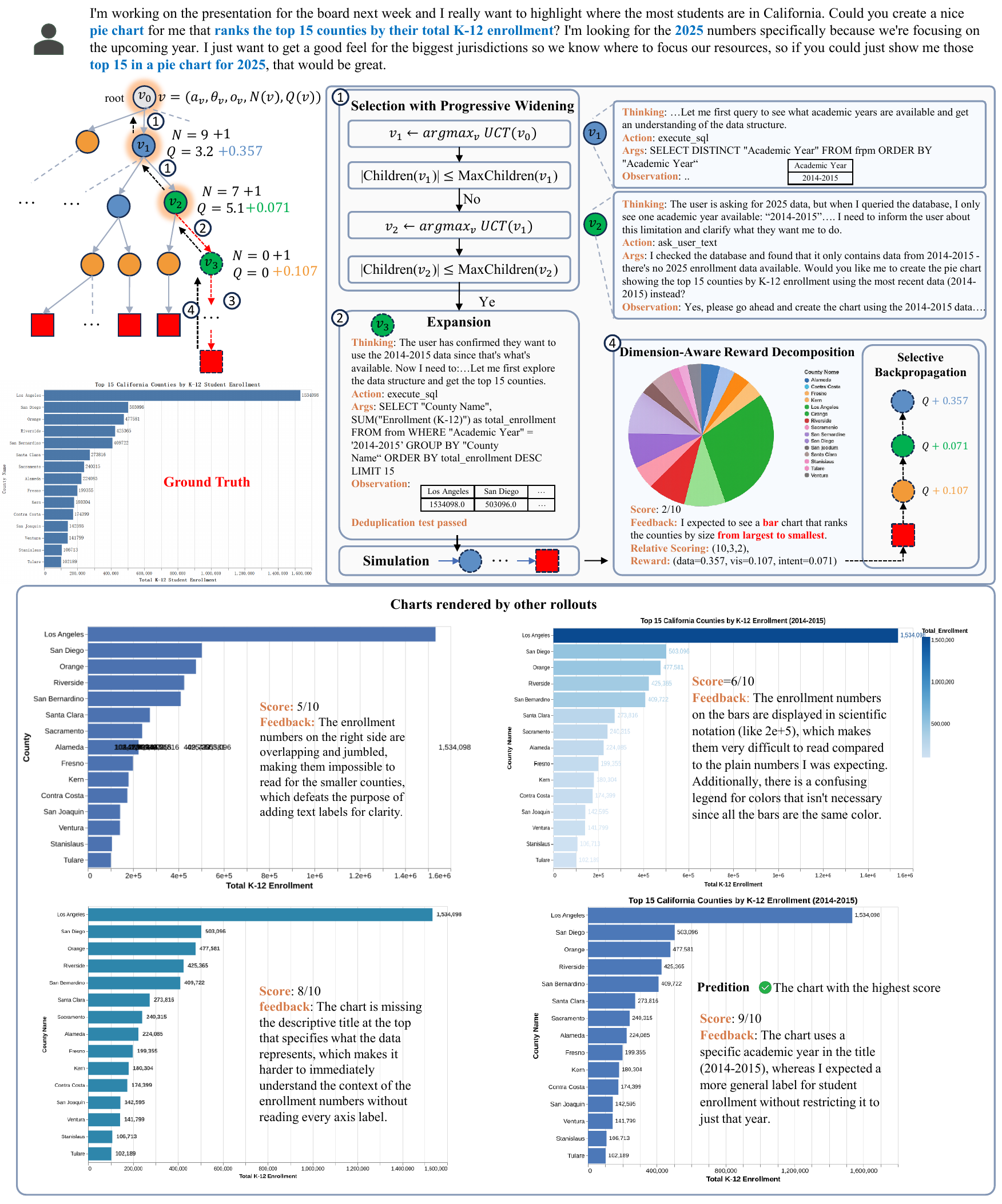}
    \caption{Case study of Vis-MCTS.}
    \label{fig:case-study}
\end{figure}

Figure~\ref{fig:case-study} shows an example Vis-MCTS run on a sample.

\textbf{Concrete example.}
At terminal node in Figure~\ref{fig:case-study}, the user provides feedback ``I expected to see a bar chart that ranks the counties by size from largest to smallest.'' with $s\!=\!2$.
The LLM assigns $(s_d, s_e, s_i) = (10, 3, 2)$;
with $(w_d, w_e, w_i) = (0.4, 0.4, 0.2)$, anchored normalization yields $\lambda = 2/(0.4\!\times\!10 + 0.4\!\times\!3 + 0.2\!\times\!2) \approx 0.357$ and $(r_d, r_e, r_i) = (0.357, 0.107, 0.071)$.
The SQL node receives $+0.357$ (rewarded for correct data), the Altair node $+0.107$ (penalized for chart-type mismatch), and the text-query node $+0.071$ (penalized for unaligned intent).
Under uniform backpropagation, all nodes would receive $+0.20$, unfairly penalizing the SQL node for an unrelated visual deficiency. In contrast, our dimension-aware reward decomposition correctly attributes $+0.357$ to the SQL node while directing the penalty to the Altair node ($+0.107$), enabling targeted credit assignment.

\subsection{Judge Reliability}
\label{app:judge-reliability}

This section provides full details for the judge-reliability study summarised in \S\ref{sec:exp-analysis} and Figure~\ref{fig:judge-reliability}.

\textbf{Protocol.}
We use a two-stage validation.
\emph{Stage~1 (LLM cross-family).} For each of the six methods in Table~\ref{tab:main-results} (under the \texttt{Qwen3.5-flash} backbone) we draw a stratified random sample of $200$ outputs, yielding $1{,}200$ samples and $5{,}617$ key-feature judgments in total. Each sample is independently re-judged by a second judge from a different model family, \texttt{GPT-5.4-mini}, using the exact same dual-judge prompts as the main paper.
\emph{Stage~2 (3-way with humans).} From the $1{,}200$ samples we further draw $300$ samples ($1{,}406$ key features) and collect annotations from human experts. The same $300$ samples are scored by all three judges, producing strictly aligned 3-way ratings on every key feature.

\textbf{Aggregate sample-level correlations.}
Table~\ref{tab:judge-sample-pearson} reports sample-level Pearson and Spearman correlations between the per-sample KF pass rates of each pair of judges, complementing the KF-level Kappa values shown in Figure~\ref{fig:judge-reliability}.
Pearson $r$ is uniformly above $0.86$ across all comparisons and perspectives (up to $0.946$ for \texttt{Qwen} vs.\ \texttt{Human} on Code), and the mean absolute difference in per-sample pass rate stays in the range $[0.05, 0.09]$, confirming that judges not only agree on individual binary KF outcomes but also on the aggregate sample-level scores used in Table~\ref{tab:main-results}.

\begin{table}[h]
\centering
\caption{Sample-level agreement on per-sample KF pass rates. $N$ = number of samples; MAE = mean absolute difference in pass rate; $r$ = Pearson, $\rho$ = Spearman.}
\label{tab:judge-sample-pearson}
\small
\setlength{\tabcolsep}{5pt}
\begin{tabular}{l c ccc ccc ccc}
\toprule
& & \multicolumn{3}{c}{\textbf{Code}} & \multicolumn{3}{c}{\textbf{Chart}} & \multicolumn{3}{c}{\textbf{Merge}} \\
\cmidrule(lr){3-5} \cmidrule(lr){6-8} \cmidrule(lr){9-11}
\textbf{Comparison} & $N$ & $r$ & $\rho$ & MAE & $r$ & $\rho$ & MAE & $r$ & $\rho$ & MAE \\
\midrule
\texttt{Qwen} vs.\ \texttt{GPT}    & 1{,}200 & 0.925 & 0.921 & 0.066 & 0.872 & 0.861 & 0.085 & 0.912 & 0.910 & 0.067 \\
\texttt{Qwen} vs.\ \texttt{GPT}    &   300   & 0.932 & 0.928 & 0.064 & 0.883 & 0.885 & 0.087 & 0.917 & 0.918 & 0.067 \\
\texttt{Qwen} vs.\ \texttt{Human}       &   300   & 0.946 & 0.945 & 0.051 & 0.867 & 0.865 & 0.090 & 0.922 & 0.918 & 0.062 \\
\texttt{GPT} vs.\ \texttt{Human}   &   300   & 0.938 & 0.936 & 0.059 & 0.928 & 0.926 & 0.052 & 0.934 & 0.931 & 0.055 \\
\bottomrule
\end{tabular}
\end{table}

\textbf{Per-method breakdown.}
Table~\ref{tab:judge-perm-plus-gpt} reports KF-level agreement, Pearson correlation, and Cohen's Kappa for \texttt{Qwen} vs.\ \texttt{GPT} on the full $1{,}200$-sample set, broken down by source method. Tables~\ref{tab:judge-perm-plus-human} and~\ref{tab:judge-perm-gpt-human} provide the corresponding per-method breakdowns for \texttt{Qwen} vs.\ \texttt{Human} and \texttt{GPT} vs.\ \texttt{Human} on the $300$-sample subset. Two observations stand out.
(i)~Agreement is consistently strong across \emph{all} methods. No method drops below $\kappa\!=\!0.63$ on any perspective, and most $(\text{method}, \text{perspective})$ cells exceed $\kappa\!=\!0.80$.
(ii)~The \texttt{ReAct} chart-perspective row is the single softest spot ($\kappa\!\approx\!0.63$ for \texttt{Qwen} vs.\ \texttt{GPT}; $\kappa\!\approx\!0.58$ for \texttt{Qwen} vs.\ \texttt{Human}); inspecting these cases, we find that \texttt{ReAct}'s outputs more often hover near the satisfied/unsatisfied boundary on visual key features (e.g., partially-correct color encodings), where any two judges naturally disagree more. Even in this worst case, the $1{,}406$-KF \texttt{GPT} vs.\ \texttt{Human} agreement on chart features is $\kappa\!=\!0.893$, indicating that the boundary itself, not the \texttt{Qwen} judge in particular, is the source of residual noise.

\begin{table}[h]
\centering
\caption{Per-method KF-level inter-judge agreement: \texttt{Qwen} vs.\ \texttt{GPT} on the $1{,}200$-sample stratified subset ($N_{\text{KF}}\!=\!5{,}617$). We report agreement rate, Pearson $r$ (= $\phi$ for binary), and Cohen's $\kappa$.}
\label{tab:judge-perm-plus-gpt}
\small
\setlength{\tabcolsep}{5pt}
\begin{tabular}{l l c ccc}
\toprule
\textbf{Method} & \textbf{Perspective} & $N_{\text{KF}}$ & \textbf{Agree} & \textbf{Pearson $r$} & \textbf{Kappa} \\
\midrule
\multirow{3}{*}{Self-Correction LLM}
 & Code  & 936 & 0.926 & 0.850 & 0.849 \\
 & Chart & 936 & 0.917 & 0.809 & 0.798 \\
 & Merge & 936 & 0.931 & 0.813 & 0.810 \\
\midrule
\multirow{3}{*}{nvAgent}
 & Code  & 940 & 0.957 & 0.864 & 0.863 \\
 & Chart & 940 & 0.944 & 0.806 & 0.795 \\
 & Merge & 940 & 0.953 & 0.807 & 0.801 \\
\midrule
\multirow{3}{*}{MultiVis-Agent}
 & Code  & 935 & 0.901 & 0.802 & 0.798 \\
 & Chart & 935 & 0.909 & 0.800 & 0.795 \\
 & Merge & 935 & 0.925 & 0.817 & 0.813 \\
\midrule
\multirow{3}{*}{ReAct}
 & Code  & 936 & 0.933 & 0.834 & 0.834 \\
 & Chart & 936 & 0.847 & 0.666 & 0.628 \\
 & Merge & 936 & 0.925 & 0.848 & 0.846 \\
\midrule
\multirow{3}{*}{Best-of-N (ReAct)}
 & Code  & 928 & 0.933 & 0.823 & 0.822 \\
 & Chart & 928 & 0.922 & 0.829 & 0.825 \\
 & Merge & 928 & 0.919 & 0.829 & 0.829 \\
\midrule
\multirow{3}{*}{Vis-MCTS}
 & Code  & 942 & 0.928 & 0.789 & 0.789 \\
 & Chart & 942 & 0.925 & 0.812 & 0.809 \\
 & Merge & 942 & 0.921 & 0.825 & 0.825 \\
\bottomrule
\end{tabular}
\end{table}

\begin{table}[h]
\centering
\caption{Per-method KF-level agreement: \texttt{Qwen} vs.\ \texttt{Human} on the $300$-sample human-annotated subset ($N_{\text{KF}}\!=\!1{,}406$).}
\label{tab:judge-perm-plus-human}
\small
\setlength{\tabcolsep}{5pt}
\begin{tabular}{l l c ccc}
\toprule
\textbf{Method} & \textbf{Perspective} & $N_{\text{KF}}$ & \textbf{Agree} & \textbf{Pearson $r$} & \textbf{Kappa} \\
\midrule
\multirow{3}{*}{Self-Correction LLM}
 & Code  & 284 & 0.954 & 0.910 & 0.907 \\
 & Chart & 284 & 0.898 & 0.790 & 0.776 \\
 & Merge & 284 & 0.940 & 0.859 & 0.853 \\
\midrule
\multirow{3}{*}{nvAgent}
 & Code  & 215 & 0.949 & 0.785 & 0.784 \\
 & Chart & 215 & 0.958 & 0.827 & 0.818 \\
 & Merge & 215 & 0.958 & 0.720 & 0.720 \\
\midrule
\multirow{3}{*}{MultiVis-Agent}
 & Code  & 301 & 0.944 & 0.882 & 0.882 \\
 & Chart & 301 & 0.910 & 0.793 & 0.784 \\
 & Merge & 301 & 0.947 & 0.860 & 0.857 \\
\midrule
\multirow{3}{*}{ReAct}
 & Code  & 194 & 0.928 & 0.811 & 0.811 \\
 & Chart & 194 & 0.830 & 0.633 & 0.583 \\
 & Merge & 194 & 0.902 & 0.800 & 0.796 \\
\midrule
\multirow{3}{*}{Best-of-N (ReAct)}
 & Code  & 236 & 0.966 & 0.892 & 0.890 \\
 & Chart & 236 & 0.907 & 0.764 & 0.762 \\
 & Merge & 236 & 0.953 & 0.893 & 0.893 \\
\midrule
\multirow{3}{*}{Vis-MCTS}
 & Code  & 176 & 0.938 & 0.797 & 0.797 \\
 & Chart & 176 & 0.943 & 0.843 & 0.836 \\
 & Merge & 176 & 0.915 & 0.789 & 0.787 \\
\bottomrule
\end{tabular}
\end{table}

\begin{table}[h]
\centering
\caption{Per-method KF-level agreement: \texttt{GPT} vs.\ \texttt{Human} on the $300$-sample human-annotated subset ($N_{\text{KF}}\!=\!1{,}406$).}
\label{tab:judge-perm-gpt-human}
\small
\setlength{\tabcolsep}{5pt}
\begin{tabular}{l l c ccc}
\toprule
\textbf{Method} & \textbf{Perspective} & $N_{\text{KF}}$ & \textbf{Agree} & \textbf{Pearson $r$} & \textbf{Kappa} \\
\midrule
\multirow{3}{*}{Self-Correction LLM}
 & Code  & 284 & 0.940 & 0.879 & 0.879 \\
 & Chart & 284 & 0.940 & 0.874 & 0.872 \\
 & Merge & 284 & 0.951 & 0.884 & 0.884 \\
\midrule
\multirow{3}{*}{nvAgent}
 & Code  & 215 & 0.930 & 0.750 & 0.736 \\
 & Chart & 215 & 0.944 & 0.793 & 0.791 \\
 & Merge & 215 & 0.935 & 0.692 & 0.662 \\
\midrule
\multirow{3}{*}{MultiVis-Agent}
 & Code  & 301 & 0.924 & 0.845 & 0.843 \\
 & Chart & 301 & 0.930 & 0.839 & 0.839 \\
 & Merge & 301 & 0.937 & 0.838 & 0.838 \\
\midrule
\multirow{3}{*}{ReAct}
 & Code  & 194 & 0.938 & 0.841 & 0.840 \\
 & Chart & 194 & 0.954 & 0.898 & 0.897 \\
 & Merge & 194 & 0.938 & 0.873 & 0.873 \\
\midrule
\multirow{3}{*}{Best-of-N (ReAct)}
 & Code  & 236 & 0.949 & 0.837 & 0.835 \\
 & Chart & 236 & 0.962 & 0.899 & 0.899 \\
 & Merge & 236 & 0.945 & 0.876 & 0.875 \\
\midrule
\multirow{3}{*}{Vis-MCTS}
 & Code  & 176 & 0.926 & 0.750 & 0.749 \\
 & Chart & 176 & 0.960 & 0.885 & 0.882 \\
 & Merge & 176 & 0.915 & 0.789 & 0.787 \\
\bottomrule
\end{tabular}
\end{table}

\begin{figure}[h]
\centering
\includegraphics[width=0.78\textwidth]{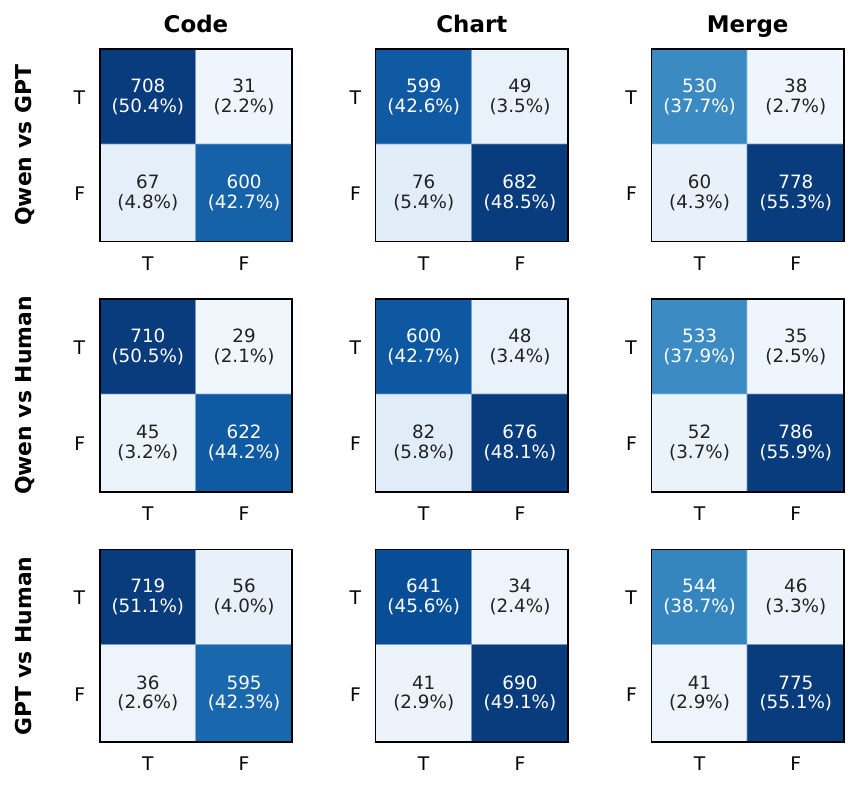}
\caption{Binary confusion matrices on the 3-way subset ($N_{\text{KF}}\!=\!1{,}406$). Each $2\!\times\!2$ panel shows the count (and percentage of $1{,}406$) of key features judged \texttt{satisfied} (T) or \texttt{not satisfied} (F) by judge $A$ (rows) versus judge $B$ (columns); rows of the grid index judge pairs and columns index perspectives. Off-diagonal cells are small and roughly symmetric across all $9$ panels, indicating no systematic over- or under-prediction by any judge.}
\label{fig:judge-confusion}
\end{figure}

\textbf{Confusion matrices.}
Figure~\ref{fig:judge-confusion} reports the binary satisfied/not-satisfied confusion matrices on the $1{,}406$-key-feature 3-way subset. Off-diagonal cells are small and symmetric. Neither judge systematically over- or under-predicts \texttt{satisfied}, and the marginal positive rates are within $\pm 2.6$ percentage points across the three judges (e.g., on Code: \texttt{Qwen} 52.6\%, \texttt{GPT} 55.1\%, \texttt{Human} 53.7\%).

\textbf{Three-way summary.}
On the $1{,}406$-KF 3-way subset, all three judges return identical labels on $90.6\%$ / $88.3\%$ / $90.3\%$ of the Code / Chart / Merge key features, with Fleiss' $\kappa_F$ of $0.874$ / $0.843$ / $0.867$ respectively. Together with the per-method breakdown above, this confirms that the rankings and gaps reported in Table~\ref{tab:main-results} are robust to the specific judge family used.

\end{document}